%% file: lm_conceptual.tex
\pdfoutput=1

\documentclass[11pt]{article}

\usepackage[preprint]{acl}

\usepackage{times}
\usepackage{latexsym}
\usepackage{expex}
\usepackage{amssymb}
\usepackage[T1]{fontenc}

\usepackage[utf8]{inputenc}

\usepackage{microtype}
\usepackage{censor}

\usepackage{inconsolata}
\usepackage{hyperref}

\usepackage{graphicx}

\title{BERT's Conceptual Cartography: Mapping the Landscapes of Meaning}

\author{Nina Haket \and Ryan Daniels \\
        University of Cambridge\\
        \texttt{\{nch35, rkd43\}@cam.ac.uk}
        }

\begin{document}
\maketitle
\begin{abstract}
Conceptual Engineers want to make words better. However, they often underestimate how varied our usage of words is. In this paper, we take the first steps in exploring the contextual nuances of words by creating \textit{conceptual landscapes} -- 2D surfaces representing the pragmatic usage of words -- that conceptual engineers can use to inform their projects. We use the spoken component of the British National Corpus and BERT to create contextualised word embeddings, and use Gaussian Mixture Models, a selection of metrics, and qualitative analysis to visualise and numerically represent lexical landscapes. Such an approach has not yet been used in the conceptual engineering literature and provides a detailed examination of how different words manifest in various contexts that is potentially useful to conceptual engineering projects. Our findings highlight the inherent complexity of conceptual engineering, revealing that each word exhibits a unique and intricate landscape. Conceptual Engineers cannot, therefore, use a one-size-fits-all approach when improving words -- a task that may be practically intractable at scale. 
\end{abstract}

\section{Introduction}
Conceptual engineering (CE) attempts to identify defects in the words that we use and tries to improve them \citep{cappelen2020}. Defects might include: incoherence, vagueness and ambiguity; moral, political, or social issues; and cognitive problems that can make our thinking deficient \citep{cappelen2018}. 

This paper aims to address the issue often overlooked by philosophers, linguists, and activists, namely that the meanings of words are highly variable and can have subtle or radical changes depending on the context of use. By utilizing computational methods from natural language processing (NLP), we can make significant contributions to conceptual engineering. This novel interdisciplinary application not only advances our conceptual frameworks but also enriches NLP research, offering new opportunities for innovation and collaboration. We argue that before engaging in any CE project, a conceptual engineer must investigate and consider the landscape of different meanings associated with a single lexical item, henceforth the \textit{conceptual landscape}. 

We follow the rapid rise of language models (LMs) used in philosophy and linguistics \cite{Aharoni2024, heersmink2024phenomenology}, and use BERT \citep{devlin2019bert} to produce embeddings for a selection of words that are frequent targets for conceptual engineers from the spoken component of the British National Corpus 2014 \citep{love2017spoken}. We use these embeddings to examine how words are clustered by context using a Gaussian Mixture Model (GMM) and compare a range of metrics. Finally, we use the GMMs to visualize and analyze the conceptual landscapes of some example words. Taken together, these methods provide a holistic view of a word's contextual usage.

We find a striking variation within and across words, highlighting the inherent complexity of CE, and the need for a tailored approach -- a task that may be intractable at scale. Moreover, our work paves the way for broader methodologies in analyzing contextual word usage and semantics, offering valuable tools for linguistic and philosophical analysis, as well as the development of more nuanced, ethical, and accurate language processing improvements.

\section{Related Work}
Conceptual Engineering is a newly evolving philosophical literature. In short, CE seeks to improve our meanings or concepts in response to a perceived defect in these meanings \cite{koch2023recent, cappelen2018, isaac2022}. For example, \citet{haslanger2000} seeks to replace a biologically-based definition of \textsc{woman} and \textsc{race} with a socially-based definition. They choose words that can make a scientific, theoretical, or social difference to push forward scientific or social progress and understanding. It is a form of activism that operates on our words and conceptual repertoires -- a normative project that looks at how we \textit{ought} to use our words.

\begin{figure*}[t]
\centering
  \includegraphics[width=1.0\textwidth]{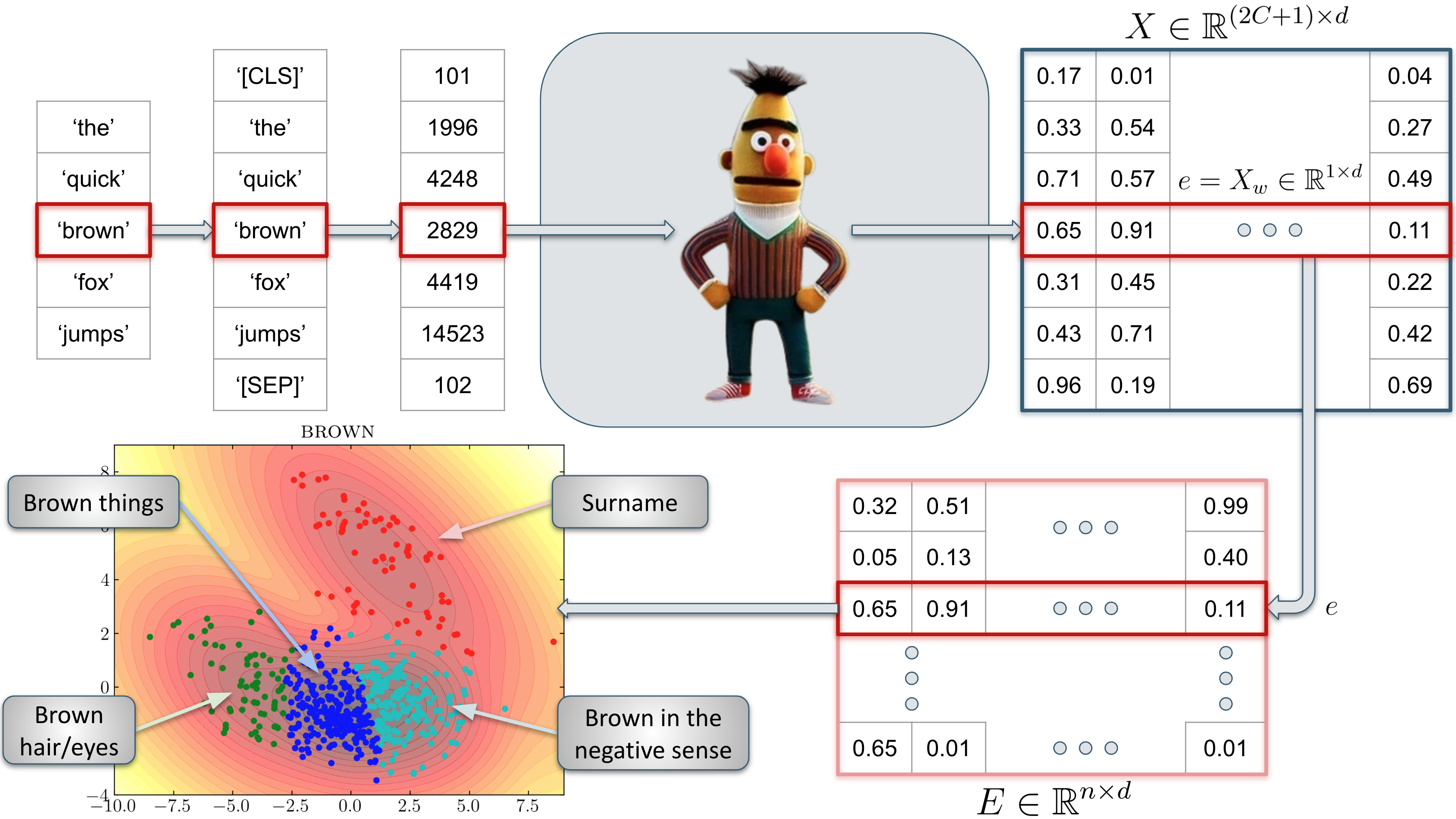}
  \caption{An example of how the target word \textsc{brown} is turned into a contextual embedding, \textit{e}. For a target word the $C$ tokens before and after $w$ are input to BERT. The final embedding $e$ for the target word is then the $w\textsuperscript{th}$ row of the embedding matrix $X$ output from the final hidden layer. A collection of embeddings taken from $n$ sentences are then collated into the matrix $E$, which is then reduced to 2D and fitted to a GMM.}
  \label{fig:bert}
\end{figure*}

For this paper, we choose to understand CE as targeting psychologically real linguistic meaning -- how real people understand and use words in discourse -- as opposed to other targets of application such as social norms or concepts \cite{haslanger2000, isaac2021a}. There are convincing reasons to do so -- this is one of the main ways we use language, and many CE projects aim for widespread change throughout a community's speech (e.g. \citealt{Cantalamessa2021} on \textsc{disability}).

The distinction between \textit{semantic meaning} and \textit{speaker meaning} is crucial to this notion; semantic meaning is the meaning independent of use, whereas speaker meaning is what a particular speaker wishes to communicate at a particular time \citep{saeed1996}. Consider the following:  "The city is busy" versus "The city is asleep". In the former \textsc{city} can be taken to communicate its semantic meaning. In the latter, taken from it cannot be the case that the \textsc{city} is asleep, since the semantic meaning of \textsc{city} is a non-sentient entity \cite{recanati2010}. Therefore, the word \textsc{city} undergoes some contextual modulation such that \textsc{city} can speaker-mean \textsc{the inhabitants of the city}. Variations in speaker meaning can happen to any word, and depend on speaker intentions and the contexts in which the word is used, and are therefore more fine-grained than the notion of `word senses' used in NLP. 

CE projects often focus on semantic meaning of words and treat speaker meaning as irrelevant \cite{cappelen2018}. However, psycholinguistic research increasingly shows that in processing and using everyday speech, we do not necessarily have to process the semantic meaning before understanding the speaker-meaning \cite{gibbs1984literal, gibbs1997pragmatics, Bezuidenhout2002-iq}. Therefore, if conceptual engineers target semantic meaning without considering speaker-meanings, their projects are unlikely to reflect how we use and process language. A focus on semantic meaning will not necessarily create the expected meaning change. If conceptual engineers wish to influence ordinary conversations and psychologically real meaning, they must consider speaker meanings when analysing, designing, and implementing their projects and frameworks \cite{pinder2020conceptual}. However, comprehensively understanding these speaker-meanings cannot be done from the armchair. Empirical methods are needed to grasp the existing scope, or conceptual landscape, of lexical usage. This paper takes a first step to operationalize this lexical variation.

Prior computational work has focused on comparisons of contextual embeddings. \citeauthor{ethayarajh2019contextual} compares the embeddings of ELMo \cite{peters-etal-2018-deep}, BERT, and GPT-2 \cite{radford2019language}, and finds that the representations become more context-specific with later layers. They also found that static word embeddings are poor replacements for contextualized embeddings.

\citeauthor{cevoli2023shades} explore the psychological understanding of ambiguity. They use BERT to generate contextual embeddings of homonymous, polysemous, and unambiguous words in a large written text corpus. They then get contextual embeddings of sentences that contain these target words. Using t-SNE, they cluster these embeddings, and calculate intra- and inter-group cosine similarities, showing that the clustering of different words vary significantly.

However, there has been surprisingly little work on applying these representations and methods to CE. \citet{kobylarz_llm} have used word embeddings to look at visualising concept change over time, and \citet{Baumgartner_magic} has examined concept drift in machine learning models when applied to game rules. Both of these approaches look at the change of words \textit{across time}. Our approach is closer to that of \citet{reuter_conspiracy} in that it uses word embeddings to visualise the \textit{current state} of a word. However, none of these approaches have focused on computationally assessing the variation in lexical meaning to inform and contribute to CE efforts. They do not quantify the variation seen within a single word, and do not comment on how distributions and patterns necessitate different approaches for engineering that particular word. 

\section{Methods}
This paper presents an application of language models to the analysis of spoken text corpora to quantify and visualise differences within and across words concerning their diversity of use. In this section, we present a brief overview of the data used, and the computational methods.

\subsection{Data}
The Spoken British National Corpus (BNC) consists of 1,251 anonymized, unscripted, face-to-face conversations recorded from 672 volunteers from a range of socioeconomic and demographic backgrounds designed to be a representative sample of the British population \cite{love2017spoken}. The conversations were collected from 2012 to 2014 in a variety of contexts, including business meetings and radio phone-ins, and therefore are representative of everyday vernacular speech. Work on spoken language is underrepresented in previous empirical work on CE. The Spoken BNC is released under the Spoken BNC2014 User Licence for non-commercial research and teaching purposes. Our usage in this paper complies with these terms.

\subsection{Contextual embeddings}
The meaning of words is intricately tied to their patterns of use within a given context, and if a word occurs in systematically different contexts, there is a distinction of meaning to be found: "You shall know a word by the company it keeps” \cite{firth-1957}. Words with similar meanings tend to appear in similar contexts and share similar distributions across various linguistic environments, and therefore share similar contextual embeddings. Contextual embeddings using BERT are therefore a good fit for quantifying speaker meaning.

BERT \cite{devlin2019bert} is perhaps one of the most widely used and understood language models in the NLP community. BERT is a transformer-based language model, trained on two simultaneous tasks: masked token prediction; and next sentence likelihood. In contrast to many popular generative LLMs, BERT is bi-directional, attending to tokens both before and after a masked token. BERT comes in a variety of sizes, and for this paper we use the 336M parameter \textit{bert-large-uncased} model from Hugging Face, and all arguments were fixed to default. BERT is a low-resource, low-complexity model, which has a proven track record on various linguistic tasks \cite{sun2019utilizing, lee2020biobert, lewis2019bart}. While newer models such as RoBERTa \cite{liu2019roberta}, T5 \cite{raffel2020exploring}, or GPT-based models offer advanced performance in certain areas, BERT's balance of performance, efficiency and simplicity make it well-suited to analyzing semantic meaning in the Spoken BNC. Additionally, we chose BERT because its efficiency allows our method to be conducted on readily available hardware in under 24 hours, making it accessible for researchers with limited computational resources. BERT is released under an Apache 2.0 license.

BERT generates \textit{contextual embeddings}. In contrast, \textit{static embeddings}, such as word2vec \cite{mikolov2013efficient} or GloVe \cite{pennington-etal-2014-glove}, learn a \textit{global} representation of a word. In other words, only one representation is learned for each word and local contextual information will be ignored. This distinction mirrors that between semantic meaning and speaker meaning. BERT generates unique embeddings for each token based on its context, giving an approximation of speaker meaning. Conversely, static models create an average representation of a word across all its uses, which abstracts away the contextual nuances, similar to how a general semantic meaning might be understood.

We generate contextual embeddings for 24 words, \textit{target tokens}, that occur within the Spoken BNC, including words commonly targeted by conceptual engineers such as \textsc{duty}, \textsc{planet}, \textsc{truth}, and \textsc{family} (for a full list see Appendix \ref{app:A}). 

We define the context window, $C$, as half the total number of tokens in the input, excluding the target token, $T_w$. For a single occurrence of the target token in the text, the total number of tokens fed into BERT is then $2C + 1$, where $T_w$ is the middle token: $[T_1, ..., T_C, T_w, T_{C+2}, ..., T_{2C+1}]$. BERT therefore takes as input a $2C + 1$ length utterance. The last layer hidden-state is taken as the output -- an embedding matrix $X \in \mathbb{R}^{(2C+1)\times d}$. The word contextual embedding is then the $w\textsuperscript{th}$ row, $e = X_w \in \mathbb{R}^{1\times d}$. For $n$ \textit{separate} occurrences of that target token within the text can be represented by the occurrence matrix $E \in \mathbb{R}^{n\times d}$.

\subsection{Conceptual landscapes}
A Gaussian Mixture Model (GMM) is a method of modelling multimodal data using a combination of $K$ unimodal distributions. We use a GMM to perform unsupervised soft clustering on the embedding matrix $E$ after dimensionality reduction with principal component analysis (PCA). We optimize $K$ and the number of principal components for each word using the Silhouette score \cite{rousseeuw1987silhoutte}. We then perform a robustness analysis using the Adjusted Rand Index (ARI) \cite{rand1971objective}. The ARI measures the similarity between two sets of cluster assignments. Practically, the ARI ranges between [0,1] with 0 indicating entirely random assignments, and 1 indicating perfect agreement between the two cluster assignments. We fix the number of principal components, then use 1000 random initializations for training the GMM. The ARI is calculated for all pairs of cluster assignments for the 1000 random initializations. We calculate the ARI with (i) 2 principal components, and (ii) the optimal number of principal components. The final labels are calculated by aggregating the results of the 1000 runs into a consensus matrix, and using hierarchical clustering on this consensus matrix.

To construct the conceptual landscapes we use the GMM fit to the first two principal components with the optimal number of clusters, and find the log-likelihood scores over a defined space (Figure \ref{fig:bert}). For the purpose of conciseness, the qualitative analysis of the conceptual landscapes are limited to the words \textsc{duty}, \textsc{planet}, and \textsc{marriage}.

\subsection{Metrics}
We use four main metrics to describe the landscapes: \textit{maximum explained variance} (MEV), \textit{self-similarity}, \textit{intra-group similarity}, and \textit{inter-group similarity}. The definitions used here closely follow those from \citet{ethayarajh2019contextual}.

\paragraph{MEV} If target token $T_w$ appears in sentence $i$ then $e_{i}$ is the corresponding embedding. The values $\sigma_1, ..., \sigma_m$ are then the first $m$ singular values of the centered occurrence matrix. The MEV is the proportion of variance explained by the first principal component, given by

\begin{equation}
    \textrm{MEV}(w) = \frac{\sigma_1^2}{\sum_i \sigma_i^2}
\end{equation}

and ranges over $[0, 1]$. In CE, MEV indicates the extent to which a contextual embedding could be replaced by a static embedding. A word with a high MEV therefore indicates a uniform consistency of word usage (for example, if the word \textsc{bark} is always used in the context of ``like a dog''). It measures the extent to which changing the semantic meaning is likely to influence the speaker meanings. 

\paragraph{Self-similarity} The self-similarity is the average cosine similarity between embedding vectors, given by

\begin{equation}
    \textrm{Sim}(w) = \frac{1}{n^2 - n}\sum_i\sum_{j\ne i} \cos{\left( e_i, e_j \right)}
\end{equation}

and ranges over $[0, 1]$. For CE, this metric gives a value of \textit{how much} variation we see within the word. A word with a high self-similarity is constrained in its diversity of usage and meaning, whereas a low self-similarity indicates high diversity in usage.

Anisotropy (the non-uniform distributions of words in embedding space) in LLM contextual embeddings is well documented \cite{ethayarajh2019contextual}. It is therefore necessary to control for anisotropy in self-similarity by taking a random sample of embeddings, and finding the total average similarity. This baseline is then subtracted from the self-similarity for each word.
\begin{figure*}[t]
  \includegraphics[width=\textwidth]{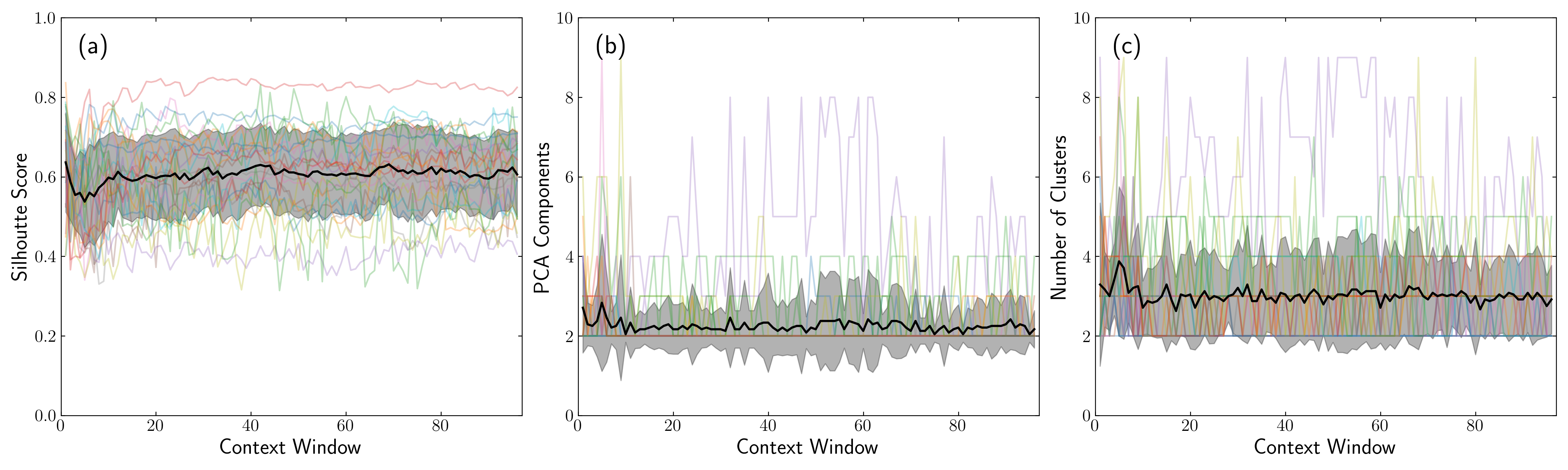}
  \caption{The Silhouette scores (a), optimal number of principal components (b), and optimal number of clusters (c) for each GMM fit to each word. Bold lines indicate averages, and shaded regions indicate the standard deviation.}
  \label{fig:context}
\end{figure*}


\begin{figure}[ht!]
  \includegraphics[width=\columnwidth]{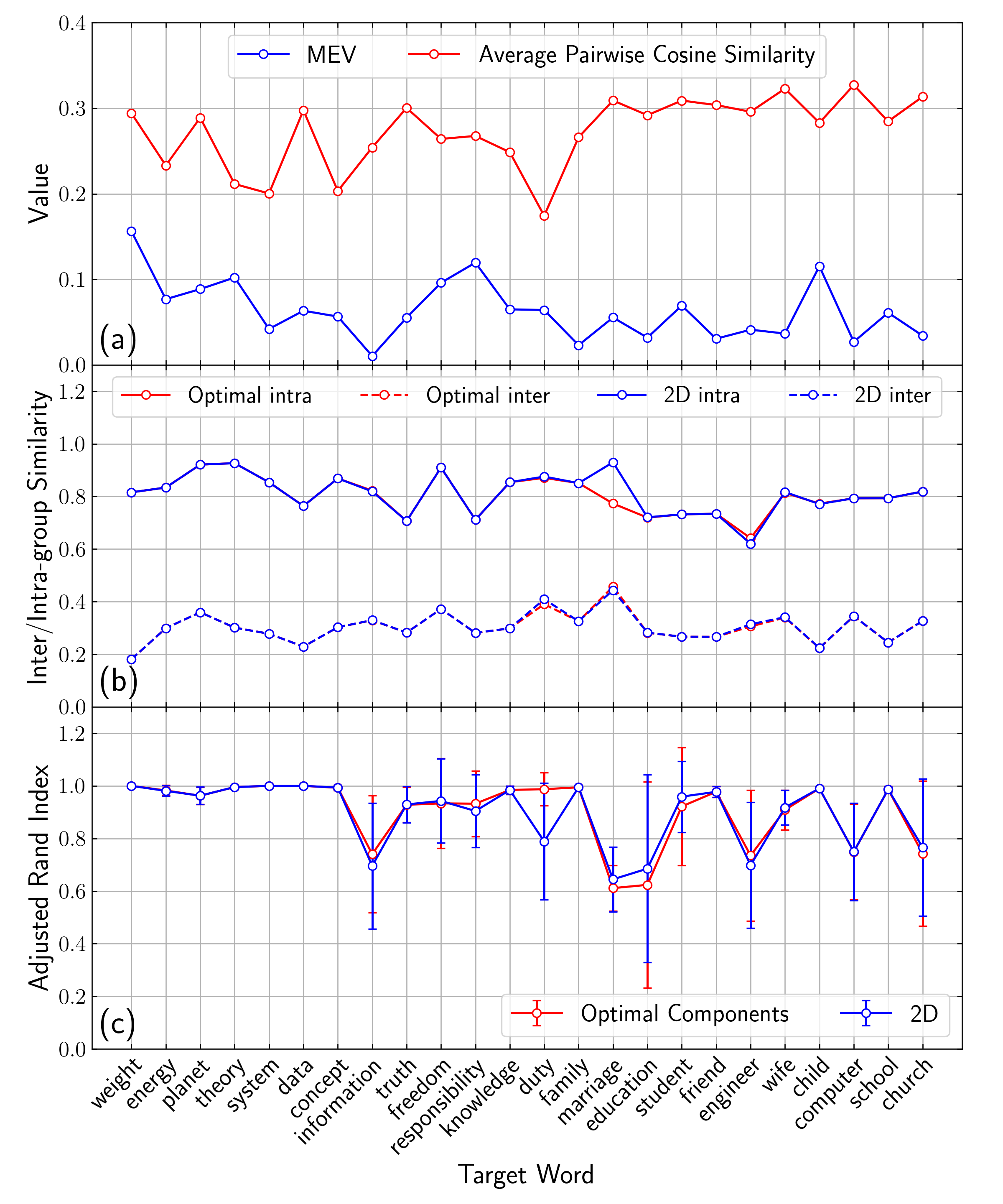}
  \caption{(a) Anisotropy-corrected self-similarity (red) and maximum explained variance (blue). (b) Intra- (solid line) and inter-group (dashed line) similarity for the optimal number of principal components (red), and for 2 principal components (blue). (c) ARI for 1000 GMMs fitted to the optimal number of principal components (red), and for 2 principal components (blue). Error bars are the standard deviations.}
  \label{fig:ari}
\end{figure}

\paragraph{Intra-group similarity} Let $e_{k,i}$ be the embedding $e_i$ assigned to label $k$ with $n_k$ members. The average intra-group similarity for $K$ groups is then

\begin{equation}
    \textrm{Intra} = \frac{1}{K} \sum_k^K \frac{1}{n_k^2 - n_k}\sum_i\sum_{j\ne i} \cos{\left( e_{k,i}, e_{k,j} \right)}
\end{equation}

For CE, this metric gives a measure of similarity within assigned contextual clusters. If the clusters contain usages which are contextually similar, this score should be high. High intra-group similarity suggests that the word is used consistently within each cluster, facilitating more precise and effective conceptual engineering interventions. This allows for targeted modifications to the word's meaning and usage, making it easier to implement changes and achieve the desired conceptual clarity.

\paragraph{Inter-group similarity} Let $e_{k,i}$ be the embedding $e_i$ assigned to label $k$, where $n_l$ are those embeddings \textit{not} assigned to label $k$. The average inter-group similarity for $K$ groups is then 

\begin{equation}
    \textrm{Inter} = \frac{1}{K} \sum_k^K \frac{1}{n_k n_l - n_l}\sum_i\sum_{j} \cos{\left( e_{k,i}, e_{l,j} \right)}
\end{equation}

For CE, this metric compares members of a single contextual cluster with members from \textit{other} contextual clusters. If the clusters are contextually different from one another, and each individual cluster contains usages which are contextually similar, this score should be low. High inter-group variation suggests more distinct boundaries between contexts, delineating specific usages, which can make CE easier to implement since it can target specific contexts without interference from others. 

\section{Results and Discussion}
\subsection{Context size}
Figure \ref{fig:context} shows the result of optimizing the GMM for (a) Silhouette scores, (b) number of principal components, and (c) number of clusters for different context window sizes for the target words. Note that the minimum value of the Silhouette score is achieved at $C = 4$, and therefore when the total number of tokens is $\sim 9$. The utterance lengths of the Spoken BNC are approximately power law distributed (see Appendix \ref{fig:linelengths}) with an average utterance length of $\sim 10$. This suggests that taking a single utterance as input to BERT may be insufficient to capture the full contextual meaning of the target word. As the total number of input tokens exceeds the average utterance length, the Silhouette score increases quickly and remains relatively steady, achieving a maximum at $C\sim 40$. 

Importantly, the average number of optimal principal components across words and context windows is $\sim 2$, and the optimal number of principal components is 2 for every word, except for \textsc{duty}, and \textsc{marriage}. For the following sections we choose a context window of 40, where the Silhouette score is at a maximum. For all subsequent analyses, the number of clusters is fixed to the optimal number of clusters for each word (for Silhouette scores, optimal principal components and optimal number of clusters for each word, see Figure \ref{fig:bestparams}).

\subsection{Cluster properties}
\begin{figure*}[t]
  \includegraphics[width=\textwidth]{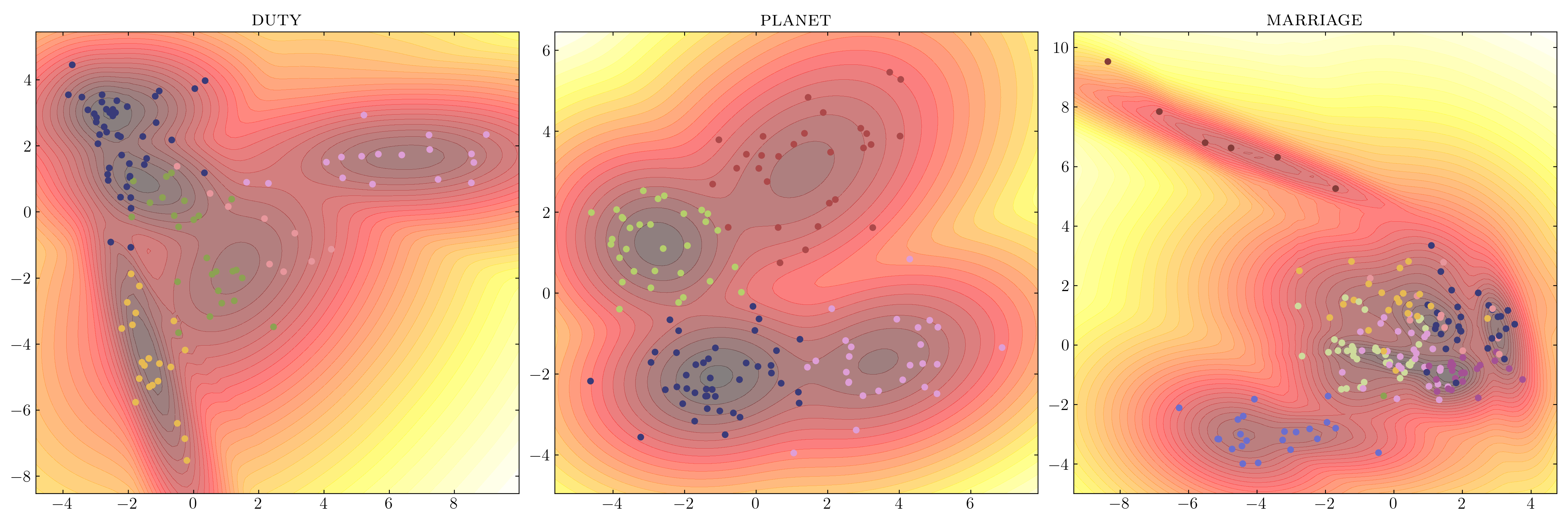}
  \caption{The conceptual landscapes generated using the negative log-likelihood of the GMM predictions in 2D for (a) \textsc{duty} with 5 clusters, (b) \textsc{theory} with 3 clusters, and (c) \textsc{planet} with 4 clusters.}
  \label{fig:landscape}
\end{figure*}

Figure \ref{fig:ari} shows the MEV scores and average self-similarities after correcting for anisotropy (a), and the intra-group similarity and inter-group similarity (b) for the target words. These results are in strong agreement with \citeauthor{ethayarajh2019contextual} that static embeddings would be poor substitutes for the contextual embeddings obtained from BERT. In addition, we also found that a control for anisotropy was not necessary when reducing dimensions.

Figure \ref{fig:ari}c shows that there is an excellent agreement between the ARI scores when using 2 principal components and when using the optimal number of components, suggesting that the 2D representations capture a substantial amount of the clustering structure found in the higher-dimensional space. Secondly, the ARI scores show significant variability across words. Words such as \textsc{weight}, \textsc{system}, and \textsc{family} have high average ARI, and low variance; words such as \textsc{information}, \textsc{education}, and \textsc{duty}, have lower average ARI and higher variance.

Words with high ARIs cluster in a consistent manner across different initializations, indicating a well-defined, stable model, and therefore a well-defined and stable conceptual landscape. The contexts are likely to be more distinct and less ambiguous. Words with lower ARIs may have more ambiguous or varied contexts, causing the clusters to overlap. Therefore, the varying levels of stability are reflective of the differences between contextual distinctions and ambiguity. The ARI scores for each word are understandably correlated with the Silhouette scores (\( r = 0.723 \), \( p < 0.0001 \)), given both metrics aim to quantify a measure of cluster quality and stability albeit from different perspectives.

\subsection{Conceptual landscapes}
Since the average number of optimal principal components is approximately 2, it is therefore reasonable to use the 2D conceptual landscape as an indicator of contextual word usage. Figure \ref{fig:landscape} shows example conceptual landscapes for the example words \textsc{duty}, \textsc{planet}, and \textsc{marriage} (for all target words and landscapes see, see Appendix A, and Figures \ref{fig:landscape1} and \ref{fig:landscape2}). 

\subsubsection{Duty}
\textsc{duty} is a frequently discussed philosophical concept and likely candidate for CE projects. The optimal ARI for \textsc{duty} is 0.99, suggesting consistent cluster choices in higher dimensions. However the 2D ARI is significantly lower at 0.79. This suggests that more information is captured in higher dimensions. This is also evident in the conceptual landscape shown in Figure \ref{fig:landscape}a -- there is significant overlap between clusters.

The MEV is 0.06, suggesting that a static embedding would be a poor substitute, and a single idealised meaning would be a poor representation of the variation of use. Therefore, targeting a single `idealised' semantic meaning for CE would fail to capture the variation found in lexical use. The self-similarity, adjusted for anisotropy is 0.17, the lowest of all words tested, and is also indicative of a high diversity of usage: there are many ways we use the word \textsc{duty}. \textsc{duty} has an intra-group similarity of 0.87, a high value compared to other words, and an inter-group similarity of 0.39. If we were to engineer \textsc{duty}, it suggests that while there are several distinct uses of the word, these uses are not well differentiated, and the project would involve a complex interweaving and picking apart of meanings and uses.

\begin{figure*}[t] 
  \includegraphics[width=\linewidth]{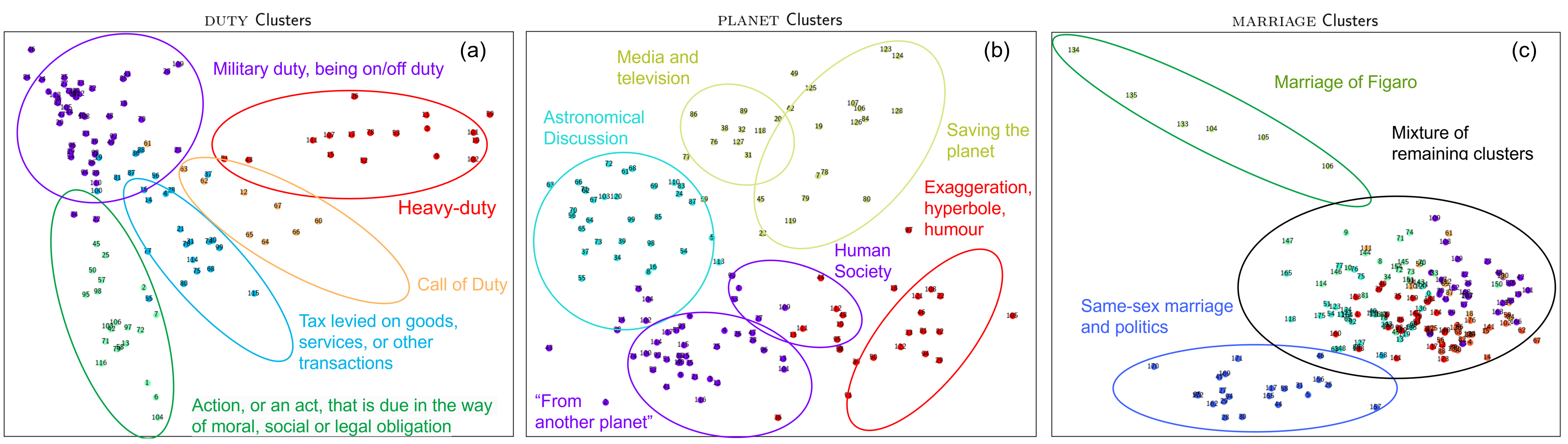}\label{fig:part3}
  \caption {Qualitative inspection of the conceptual landscapes for (a) \textsc{duty}, (b) \textsc{planet} with 4 clusters, and (c) \textsc{marriage} with 9 clusters.}
  \label{fig:part1}
\end{figure*}

To expand on these findings, we performed a qualitative analysis on the conceptual landscape of \textsc{duty} (all contexts are available in the Supplementary Materials). We found that the clusters correspond to 5 usages (Figure \ref{fig:part1}a): 1) being on/off duty, 2) taxation on goods and services, 3) moral, social or legal obligations, 4) the phrase `call of duty', and 5) the phrase `heavy duty'. Conceptual Engineers who wish to engineer \textsc{duty} would likely target the use of \textsc{duty} as referring to moral, social or legal obligation. In this case, since the clusters are well-defined but display overlap, it seems difficult to target a single meaning. However, several of the different uses pertain to set phrases that \textsc{duty} features in ("call of duty", "on/off duty", "heavy-duty", "duty-free") which may have become opaque to language users; they process the whole phrase without looking at the individual components  \cite{Vega-Moreno2007-du}. As such, they may not interfere with modification of {duty} as referring to moral, social or legal obligation. 


\subsubsection{Planet}
\textsc{planet} has dominated the CE literature \cite{LandesManuscript-LANCRI-3} since the \citeauthor{iau2006definition} redefined the term so as to exclude Pluto. The ARI for \textsc{planet} is 0.96, suggesting consistent cluster assignments. The MEV is 0.09, highlighting the insufficiency of a single static embedding to capture the usage. The self-similarity, adjusted for anisotropy, is 0.29, amongst the higher values for the target words, and yet still indicates a high diversity of usage. \textsc{planet} has an intra-group similarity of 0.92, indicating high coherence within clusters. This suggests that conceptual engineers have clear usages to target, but a high inter-group similarity of 0.36, may suggest a lack of differentiation. 

We also find more distinct uses than there are clusters (Figures \ref{fig:landscape}b and \ref{fig:part1}b). While the GMM identified 4 clusters, analysis of the respective contexts identified 6: 1) "from another planet", 2) Astronomical discussion, 3) saving the planet, 4) media and television, 5) human society, and 6) exaggeration and hyperbole. This highlights the need for qualitative analysis. Furthermore, there are fewer entrenched phrases, limited perhaps to "from another planet", suggesting that the different meanings and usages are much more interwoven than in \textsc{duty}. For example, \textsc{planet} can be said to be modulated to refer to aspects of human society through the process of metonymy or to be used in cases of hyperbole (in which the effect is created through the difference between the semantic and speaker meanings, e.g. \citealt{colston2020}). As such, we can suggest \textsc{planet} would be hard to engineer: choosing a single meaning to engineer could have far-reaching and unintended consequences.

\subsubsection{Marriage}
Along with philosophical and scientific uses of CE, there is a rise in the use of CE on social concepts, such as \textsc{marriage}.

The optimal ARI for \textsc{marriage} is 0.61 (2D = 0.64), much lower than the other examples, suggesting potential ambiguity between clusters. The MEV is 0.06, and the anistropy adjusted self-similarity is 0.31, which is high compared to other words, suggesting less variation in use. \textsc{marriage} has an intra-group similarity of 0.77, and inter-group similarity of 0.46, which is higher than any other word. These metrics suggest poor differentiation between its 9 clusters. 

Looking at the landscapes (Figure \ref{fig:landscape}c) and the contexts (Figure \ref{fig:part1}c), this is indeed what we find. There are two distinct clusters pertaining to discussions of same-sex marriage and references to the opera \textit{The Marriage of Figaro}. The remaining 7 are not highly differentiated. \textsc{marriage} is one of two words for which 2 dimensions suboptimal. We opted here for 2D to ensure consistency, but the interweaving of several clusters may be less extreme in higher dimensions. Regardless, we see here that despite being assigned different clusters, these remaining 7 clusters display similar contextual properties so as to be concentrated in a small area of space. Conceptual engineers suggest that \textsc{marriage} should be engineered to include gay marriage. Within the Spoken BNC, there are contextual differences between different types of marriage: uses of "same-sex marriage" occur in distinct contexts to other uses of "marriage". This may coincide with the legalisation of gay marriage in the UK in 2013. To infer whether they are now on equal footing, we need more modern spoken corpora to see whether the contexts have merged. 

\section{Conclusion}
In this paper, we introduce an approach to conceptual engineering by constructing \textit{conceptual landscapes} through the use of BERT embeddings and Gaussian Mixture Models. 

We consider multiple metrics obtained from the embeddings and GMMs, and also perform qualitative analyses of conceptual landscapes, to gain a nuanced understanding of word usage in spoken English. It is important to note that each metric offers a different insight: the MEV draws comparisons with static embeddings; the similarity measures consider the overall diversity in usage, and the coherence and distinctness of the contextual clusters; the analysis of GMM cluster quality via the Silhouette scores and ARI examines the stability of the conceptual landscapes.

A key finding of this work is the holistic nature of evaluating contextual word usage. Our analysis reveals that words form complex and variable landscapes that is beneficial to both CE and NLP applications. Some words reveal well-defined, but overlapping clusters, requiring a targeted approach to CE. Others display diffuse and highly unstable landscapes, even in higher dimensions, suggesting that CE efforts must consider the broader usage across multiple connected contexts rather than a single idealised semantic meaning.

This paper highlights the inherent challenges of CE, but also offers a methodology for addressing some of these challenges using NLP methods. We hope that by taking a `multi-modal' approach of diverse metrics and visualization, we can provide a framework that can inform CE efforts, and help identify effective strategies for modifying word meanings. We can develop a deeper understanding of the nuances in word usage and context, ultimately contributing to more sophisticated models of language and meaning.

Combining NLP methods with Conceptual Engineering offers interesting future avenues. Methods can be developed to reflect current linguistic usage and actively shape and improve our conceptual landscapes to aid in downstream tasks like sentiment analysis and bias detection. These methods would better capture nuanced and evolving meanings, making NLP applications more responsive to cultural and contextual changes. NLP contribution to this fast-growing philosophical field could aid in better achieving the social and theoretical reform CE wants to bring about. Finally, we hope that this work encourages the use of analytical methods to quantify and compare conceptual landscapes, and further encourages interdisciplinary collaboration on CE.

\section{Limitations}
BERT is pretrained on \textsc{BookCorpus} \cite{zhu2015aligning} and English \textsc{Wikipedia} \cite{devlin2019bert}, which may introduce biases reflective of these contexts into our analysis. By adjusting for anisotropy, we mitigate some of these biases. However, this is not a complete solution. Future work should explore other models and fine-tune on more diverse datasets. In addition, the Spoken BNC includes speech from British individuals over a limited time period, which may not reflect contemporary language use and perspectives, and does not encompass linguistic data from other countries.

While 2D projections are useful for visualizing and comparing word contexts, there are instances where higher-dimensional embeddings (e.g., for \textsc{marriage}) provide a clearer representation of semantic differences. This highlights a limitation of our current approach, as projecting down to 2D may obscure important nuances in some cases. Future work should explore the use of higher-dimensional embeddings more extensively to ensure a consistent and comprehensive analysis of word meanings. PCA is also a linear transformation. There are many nonlinear dimensionality reduction techniques that we did not explore here such as t-SNE and UMAP that may aid in cluster visualizations.

A natural critique of corpus linguistics is that there is an “inevitable focus on surface forms in corpus work” \cite{Adel2010-}, suggesting we may be left empirically impoverished when analyzing data as a second-hand observer. We acknowledge such a limitation, while also arguing that the nature of CE requires an approach such as ours. CE is distinctive in that it is practical, applied, and “out in the world”. If we want our CE theory or project to be successful, we need information about how people use the words we wish to engineer.

\section{Ethical Considerations}
Ethical considerations for this work span two domains: the use of language models; and the implications of modifying the ties between language and concepts, and the associated societal impacts.

\subsection{Use of language models}
\paragraph{Cultural and language bias} BERT's training data inherently contain biases and prejudices that were present within society at the time of collection. In particular, it has been found that \textsc{BookCorpus} contains problematic content, and skewed religious representation \cite{bandy2021addressing}. These biases can manifest in the outputs of any downstream tasks. The framework presented here can potentially be used to identify biases present in text corpora and training data.

The language used in BERT's training data is predominantly English, and many other global demographics and culturally unique linguistic patterns are underrepresented. This could lead to inequitable outcomes in downstream applications of BERT. A possible mitigation is the further fine-tuning of BERT on language- or culture-specific data.

\paragraph{Environmental impact} As LLMs become more pervasive, the toll on energy consumption increases. Our attempts to mitigate this issue are a driving factor behind the use of BERT, a lightweight model, easily run on consumer hardware, consuming relatively little power.

\paragraph{Privacy and copyright concerns}
Much training content is scraped in an unsupervised fashion. This could lead to the inclusion of private information, or of content used without the consent of the creator, raising concerns about privacy and the ethical use of data. Since BERT was trained on \textsc{BookCorpus} and English \textsc{Wikipedia}, privacy concerns are mostly mitigated. However, \textsc{BookCorpus} contains data from authors whose consent was not sought.

\subsection{Conceptual engineering}
CE aims to modify word meanings, which can appear prescriptive and dismissive of natural language evolution. There are deep connections between language, culture and identity. If these changes are not inclusive of a diverse array of perspectives, CE can potentially do harm to the communities they wish to help. 

There are also wider societal and political implications with changing word meanings that could deepen existing animosity or conflict between groups. We stress the importance of a diversity of views and broader societal ties to language. We hope that this paper and methodology will provide another tool for conceptual engineers to understand current lexical usage, and help mitigate unwanted implications. We express no opinions on what words \textit{should} mean and make no judgements regarding the lexical items we investigated, but simply provide data to describe words in their current state.

\section*{Acknowledgments}
This work was performed using resources provided by the Cambridge Service for Data Driven Discovery (CSD3) operated by the University of Cambridge Research Computing Service (www.csd3.cam.ac.uk), provided by Dell EMC and Intel using Tier-2 funding from the Engineering and Physical Sciences Research Council (capital grant EP/T022159/1), and DiRAC funding from the Science and Technology Facilities Council (www.dirac.ac.uk). RKD acknowledges funding from the Accelerate Programme for Scientific Discovery Research Fellowship. NH acknowledges funding from the Cambridge Trust's International Scholarship. The funders had no role in study design, data collection and analysis, decision to publish, or preparation of the manuscript. The views expressed are those of the authors and not necessarily those of the funders. The authors thank Aditya Ravuri and Catherine Breslin for useful discussions and feedback.

\bibliography{custom}

\appendix

\include{appendix}

\include{supplementary}

\end{document}

%% file: appendix.tex
\onecolumn

\section{Full list of tested words} \label{app:A}
\begin{itemize}
 \setlength\itemsep{0em}
    \item weight
    \item energy
    \item planet
    \item theory
    \item system
    \item data
    \item concept
    \item information
    \item truth
    \item freedom
    \item responsibility
    \item knowledge
    \item duty
    \item family
    \item marriage
    \item education
    \item student
    \item friend
    \item engineer
    \item wife
    \item child
    \item computer
    \item school
    \item church
\end{itemize}

\newpage
\counterwithin{figure}{section}
\section{Conceptual Landscapes}
\label{sec:appendix}
 \begin{figure*}[h!]
  \includegraphics[width=\textwidth]{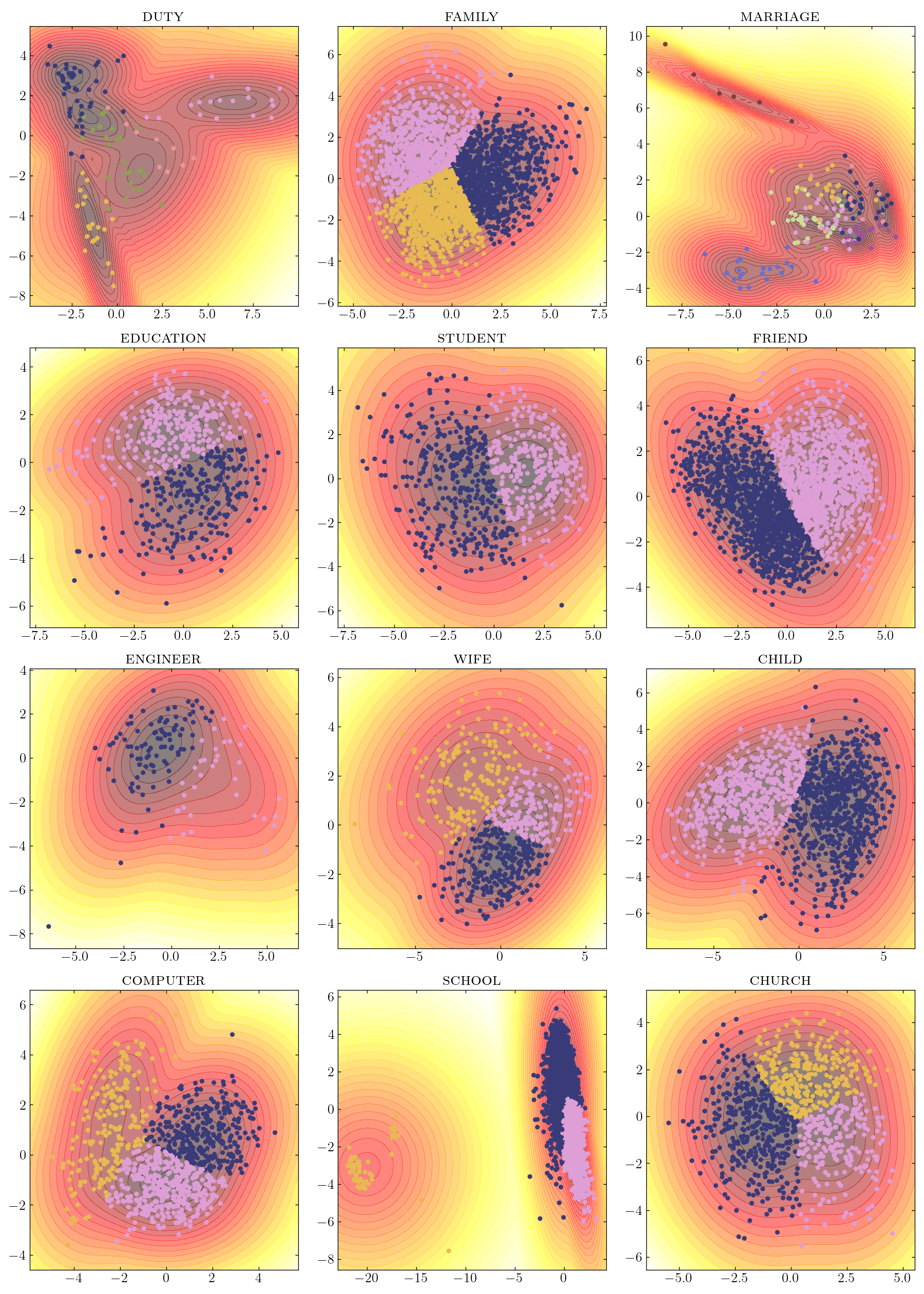}
  \caption{The conceptual landscapes generated using the negative log-likelihood of the GMM predictions for \textsc{duty} through to \textsc{church}.}
  \label{fig:landscape1}
\end{figure*}

\begin{figure*}[!ht]
  \includegraphics[width=\textwidth]{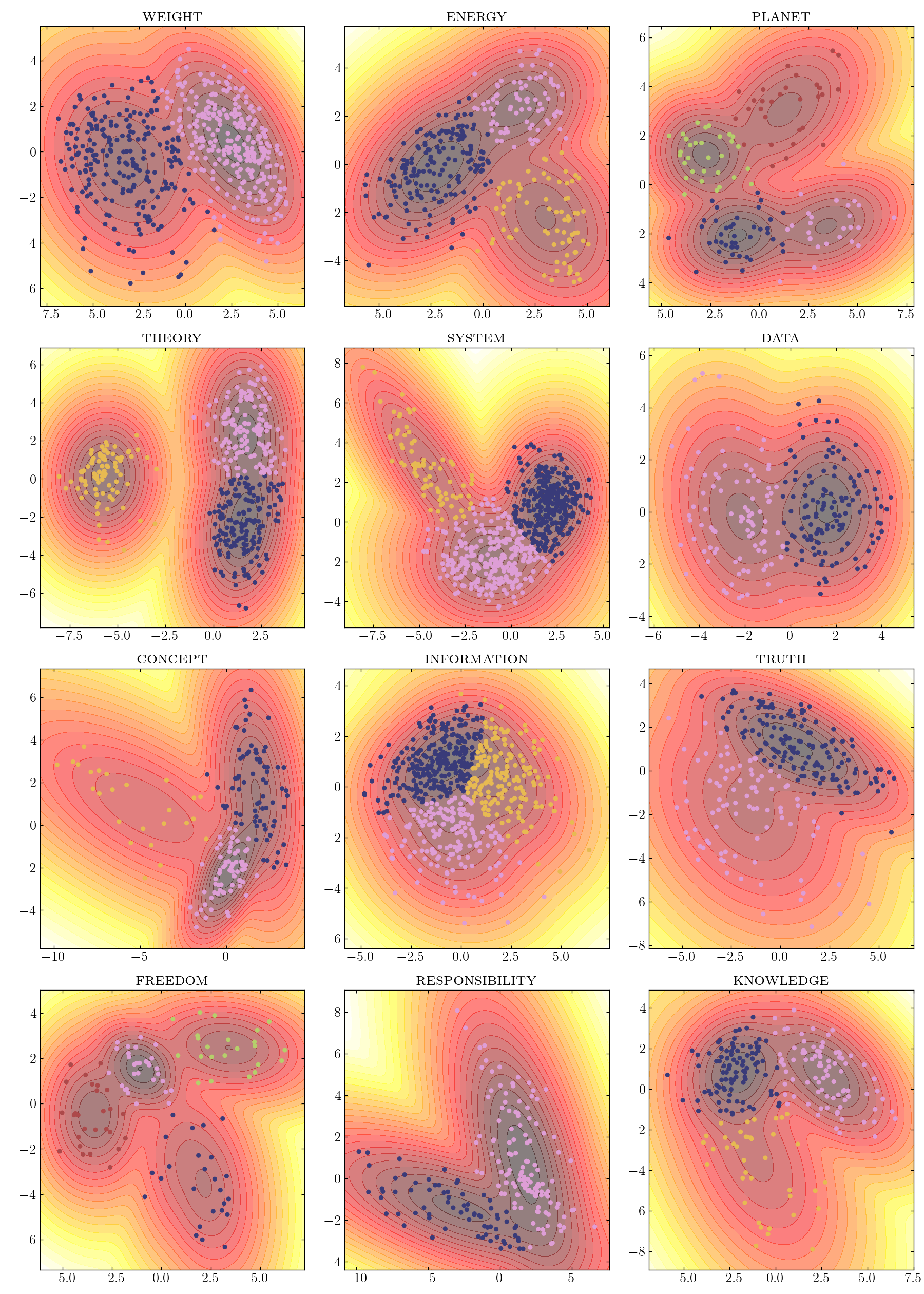}
  \caption{The conceptual landscapes generated using the negative log-likelihood of the GMM predictions for \textsc{weight} through to \textsc{knowledge}.}
  \label{fig:landscape2}
\end{figure*}

\section{Parameter Optimization}

\begin{figure*}[h!]
\centering
  \includegraphics[width=0.75\textwidth]{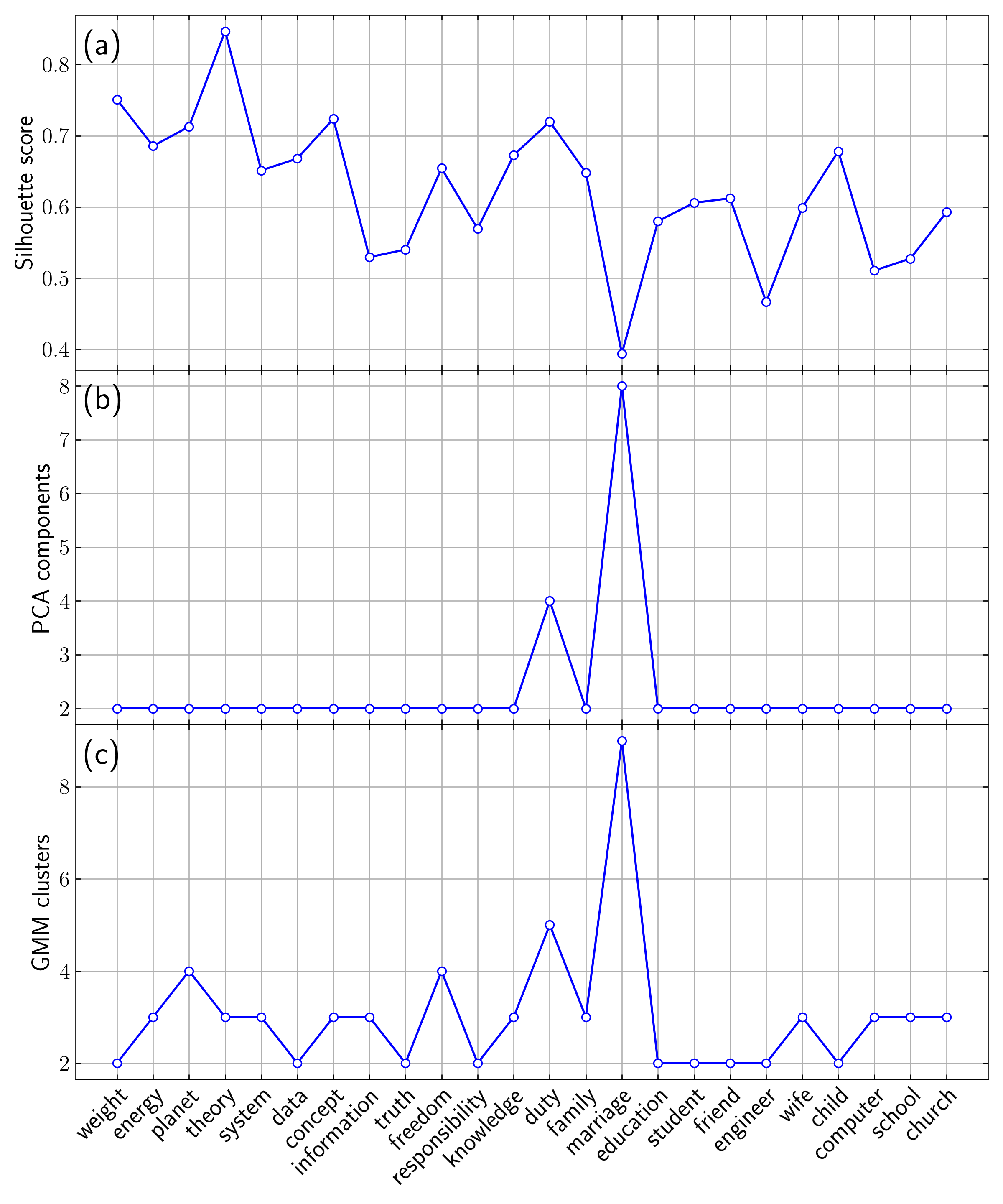}
  \caption{The resulting best Silhouette score (a) from hyperparameter optimization - the optimal number of principal components (b), and the optimal number of clusters (c). Note that the Silhouette scores correlate heavily with the ARI scores.}
  \label{fig:bestparams}
\end{figure*}

\begin{figure*}[h!]
 \centering
   \includegraphics[width=0.5\textwidth]{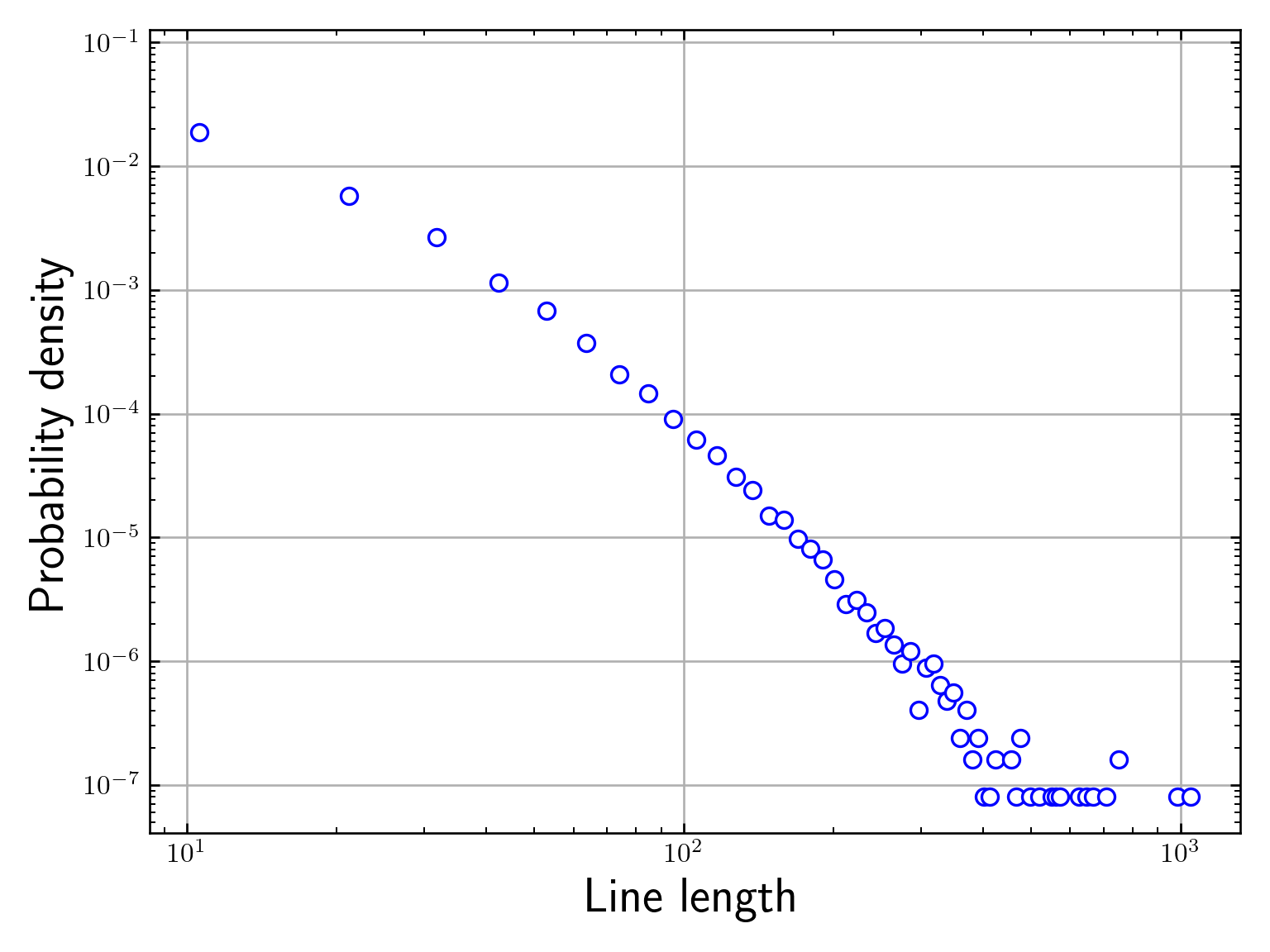}
   \caption{The resulting best Silhouette score (a) from hyperparameter optimization - the optimal number of principal components (b), and the optimal number of clusters (c). Note that the Silhouette scores correlate heavily with the ARI scores.}
   \label{fig:linelengths}
 \end{figure*}

%% file: supplementary.tex
\onecolumn
\newgeometry{left=1.5cm, right =1.5cm,bottom=3cm, top= 2cm}

\section*{Supplementary Material}

\subsection{All contexts for \textsc{duty}}

Cluster 1
\begin{itemize}  \setlength\itemsep{0em}
 \setlength\itemsep{0em}
    \item 0. i haven't passed her so i don't know but cos she cleans up at school now she's like oh does she? the cleaning supervisor so she comes down to do duty mm goes back and then around three o'clock she's back for cleaning till about six oh right so she's not about much at the mo they've got a dog now so you
    \item 5. in his skills sets than running this and who was the erm this the erm graduate was going to be taking over until january mm erm but he curtailed his erm his tour of duty to get one in london cos i was thinking he said about his girlfriend but i was thinking you didn't want all that responsibility on your shoulders mm and what the mindset seems to be now
    \item 8. so no right time to go to bed it's gonna be a big day tomorrow no? yeah try and get some sleep in the bank especially if i'm going to have to be on duty if you boys are gonna have a table tennis table tennis tournament or dart tournament or like can't play sports or i bet he can play a bit of he's not masculine enough t bet he
    \item 18. theatre and then because i think her her father - in - law must've erm been involved in some sort of anaesthetic when she went to theatre because he was the only one on duty so he came so obviously then he'd seen the baby not her choice then he told his wife obviously and they told all their side of the family before she'd even had chance to hold the
    \item 20. s some weird weirdy people about why hasn't there something been done about it mm oh strange strange and yes i mean yesterday i was out on the field watching them lunch time on lunch time duty outside this lad got upset because he got hit by the ball by the football he was standing behind the goals talking oh right and he got hit three times and the third time he picked the ball and he
    \item 22. doing enough mm should have as well really but it didn't really happen that way it's basically over now anyway i never realised that like it's like every single like almost every single duty is like can i have cover please? can someone cover me please? she like doesn't it's so weird i never noticed it before but yeah she does she never does her duties all the
    \item 23. no you're giving it to him there you go er er er get a green get a green wedge yeah that was a weird question it was a weird question i i would have gone call of duty cos i would have forgotten doom was actually a thing er er right now er sorry i'm just decimating you except i'm not because i'm not killing every tenth version of you
    \item 24. other parents are going? oh have i don't it right? well done bloody hell considering that's just a have you got your glasses? no cos no when i went to lunch duty i thought well shall i take a pair of glasses with me? didn't bother? just in case and now you can't see but coming cos i walked straight down you see oh okay
    \item 27. i i'd slept in a bi on saturday and matt got up with eh girls and then on sunday which was mother's day at then to six when my girls woke up i was on duty so that didn't seen quite right well i'm just definitely looking forward to a time when eh my son is a teenager and he eh from what i heard they sleep most of the time yeah
    \item 28. t know something about a police officer punching a woman in the face oh i hate police mm me too a lesbian couple saw their dream vacation turn into a nightmare when they were harassed by an off - duty police officer oh my god when the pair were sharing affection in the supermarket brilliant brilliant jesus christ yes i mean can we just not be a bit more mature like there's kind of a war going on
    \item 29. ' t yeah she's really really annoying i know i'm not we've got oh yeah we've got a court meeting on wednesday oh we've also got prefect oh yeah duty on that break oh yeah i need to do some of my lawyer stuff have you done the questions? some of them? you've not done any of them? i didn't do any last
    \item 32. that weird the hot water isn't that weir mm have you been working today? no no not today no oh no what have you been up to today then? i've been to lunch duty oh have you? haven't done a lot this morning actually i had a horrible sleep last night violent headache oh no yeah a really bad headache and i got you look a bit tired actually yeah i
    \item 33. it'll be crunchy on the w mm er y well you can have them did you have lunch today? no oh no i'm fine well i forgot you'd come from lunch duty i'm fine no i'm fine do you want some malt loaf? no no no honestly i'm fine thank you thank you i think that is an alarm isn't it?
    \item 34. spend a bit of time there and things but mm like when she gets it like in her mind it's like that's yeah kind of it then isn't it? oh oh well duty achieved yeah exactly but you know seen her done but mm just like mm could've been here little bit longer sort of like planned to you know have lunch or something mm mm no just mince pie
    \item 35. liners because they wait so long to go for a pee cos they never get time to go no you know so you you go hours and hours and hours i know that i have gone on duty and i've been there at half past six in the morning because of the way the buses have been running down the road i've been there at half past six no at seven to start at
    \item 39. is there not enough right for? thank you i've i've asked uncle he doesn't want to no no no i would rather have sit back you can be on timer and dice duty oh dear oh i need to wipe all the stuff off er that that's far too er oh the on there it's difficult well have the box lid and then you can shake the well dad
    \item 40. s difficult well have the box lid and then you can shake the well dad's got the box guacamole who wants the box lid? uncle he's on he's on duty he's got the yeah right it you mean i've got to actually stay awake? yes that's the erm when you roll the dice and then oh whichever word you get that '
    \item 41. mention here because in the in the paper em i didn't want to give my name for obvious reasons mm mm cos i worked in the city centre but i was just named as an off duty security guard where was this? just by the side of in front of shit do you know what's mental? i was saying to these earlier right when i went to see the navy person i '
    \item 44. the back there to the back door can see oh so we need to be more erm a little bit of decoration put a red one round here then er er excuse me i'm on red duty you just do pigging silver alright chill do your own colour scheme oh piddle you're not doing a very good job are you? i'm just saying today? sorry? what page is
    \item 46. oh it's back on now i don't know it must turn itself off afterwards well anyway yeah so yeah so has told me today mm when i went to lunch duty that erm apparently's found a lump oh in her breast oh no they've d they've done a blood test and they're sending her straight for a erm she's
    \item 47. but i just thought i'd ask on the off chance she said yeah come round at nine oh that was handy so so then now that leaves me a free day tomorrow with the exception of lunch duty to do as well oh are you on lunch duty as well? cos my plan was i was gonna do sort of either downstairs or upstairs today and then the opposite tomorrow mm and now i've
    \item 48. off chance she said yeah come round at nine oh that was handy so so then now that leaves me a free day tomorrow with the exception of lunch duty to do as well oh are you on lunch duty as well? cos my plan was i was gonna do sort of either downstairs or upstairs today and then the opposite tomorrow mm and now i've got to do it all tomorrow do you want to
    \item 49. ve got to do it all tomorrow do you want to come and do mine? can you fit that in tomorrow too? easy - peasy you could like do yours in the morning go to lunch duty do mine in the afternoon yours in the afternoon go on you know you want to do you reckon i could? oh your house is too big now i know it is too big to do it in one
    \item 54. over the journey yes yes and the stress exactly yes i mean of course you will and that's right so that's well it might be a good deal well if they want me naturally when duty calls make sure you leave your international roaming number otherwise you know it'll be the cleft stick i hardly ever going down i hardly ever switch my phone on when i go abroad on holiday yeah right
    \item 69.\# \#s i'm full of coal er p not puzzle i guess puzzle will do yeah to work out yeah cos we got what where the hell our pitch was cos we no one was on duty for one reason or ano i think it was hard to find in the first place er and then like we maybe drove around the whole kind of area like the state park and we found the main camps
    \item 70. other ambulance that came along mm and they got got in and we were parked behind these ambulances and they waited for these two this these people to come cos they were the only medics on duty that could give this injection she needed oh or somebody or or maybe they'd got it on their ambulance whatever i can't you know i guess mm and and they had to wait and these guys
    \item 73. slags? or well yes i did but i thought it was perfectly alright well no it actually wasn't perfectly alright cos they've now made a complaint against you and you were on duty and you were working for us yes and you can't just go to conferences and turn up in somebody's bedroom at one o'clock in the morning drunk and accuse somebody of being a sl
    \item 82. like one of the other teachers from my school like i had to keep my class in cos they were being a pain in the backside they were being horrible but i was supposed to be on break duty and i walked out of my classroom um and um one of the year six teachers was there and he and i said to him i was like please can you just go and outside and cover my break duty
    \item 83. and i walked out of my classroom um and um one of the year six teachers was there and he and i said to him i was like please can you just go and outside and cover my break duty cos i've gotta stay in with my class cos they're being horrible and he didn't bat an eye he just went yeah of course turned round got his coat and walked out
    \item 84. stay in with my class cos they're being horrible and he didn't bat an eye he just went yeah of course turned round got his coat and walked out and went and stood on duty if that had been me and someone had said that to me i'd have been like what really? for the whole thing? what so i don't get a cup of coffee? what so
    \item 85. i would be about oh gosh twelve? thirteen? was he involved in combat? yeah all the time? not well i don't know about all the time but there he was on convoy duty and because he was playing in a football match and he broke his toe so he couldn't sail with the convoy and his boat sunk oh and that happened twice lucky blimey the second time he
    \item 86. talked about what she had fought and then the po then the fox became a police officer and the best bit was when they had when they got when the rhinos said they had erm parking ticket duty and they were then he said and and then he told them what they needed to do and they were driving mm and then a really really fast car zoomed past them and then they said they put the
    \item 89. i i do remember yes it was the same i can remember queuing up for the the meals but then i remember you you did as a prefect right you know a sixth former you could do dinner duty oh so you you were in mixed? yes all the time you were in mixed and did you have your dinners in the boys'school hall? well only when i was er in the junior school in
    \item 90. fact? erm no i mean i picked it up but i didn't learn it no no no no but of course i was working within as a volunteer for the british council in my off duty times and it was basically english mm and sadly in my sessions for the foreign did you do singing as well at that time? no i'd given up singing at that time i picked it up again
    \item 91. i suppose that erm there's probably more peace time enough jobs and are based in this country of administrative nature that that if you have a child you could be you know sort of confined to duty at some base either here or in germany or that you wouldn't have to be no but again it could mean again that your male colleagues are taking an unfair portion of work to be done it '
    \item 92. you take that to art with you like look at my pencil case guy i challenge you to read it as quick as you can have we got english tomorrow? no no wednesday we've got prefect duty tomorrow oh yeah at break don't forget i will you will i know you i will forget i've got what have i got before break? geography and you've got re re wait for
    \item 93. like not very nice? yeah she likes doing things properly and it's like yeah oh what have you done now? i'm not looking forward to it i'm looking forward to prefect duty yeah i hope someone comes in so we can go what are you doing? yeah i think like well i like the block i'm glad we managed to switch to that yeah so am i because the
    \item 94. ? who's gonna sit up there to eat their lunch? why don't you just sit in the common room or the hall or something? yeah i wish i'd done the hall duty i'm so jealous of i could go right one two three four go you stop yeah i'd love to just st stand there and split up groups of friends yeah like leave one person left mm
    \item 96. explanation but she seems to follow it generally at the time yeah which helps yeah that is good yeah uh so we had the audiologist and then on thursday i had a phone call late morning from the duty rota organizer from link mm who wanted a meeting because it appears that some a couple of the carers have said they can't manage granny mm with in terms of the lifting without straining themselves oh
    \item 105. right some surgeries have started to break that down as well now they're starting to do different types so you can have a telephone consultation mhm you can have a quick five minute with a duty doctor which is one thing only mhm cos what a lot of people do is they come in and it's like well actually yeah i've got this pain in my foot and erm
    \item 108. but the the remainder are they the remainder? yeah no it's there's more apples but this is the remainder of the stuff i've already peeled so i might get you on peeling duty in a minute cos i'm going to need more that this probably are you? i can see there's still quite a lot of peel on those apples darling no there's not there
    \item 109. yeah yeah take my bike take my bike take all my stuff and then erm he's still he's still watching him watch him start to drive off and then er a police off - duty policeman gets out of his car and just goes bang bang bang bang bang shoots the guy that stole the bike and then he just runs after his bike and just like shit there's a guy like dying
    \item 110. mhm i drink from a sippy cup yeah that's true you do do that um i can't remember the my my alphabet in order very well hence i was telling you about dictionary duty really pr pr really primary stuff like really kiddy kiddy things mm i just never never seem to have got rid of it's isn't it how weird's that? i dunno
    \item 112. poor was awake i was bored shitless we were on duty oh no he watched every episode of gossip girl on the tv on the plane yeah what else did we watch? oh crazy stupid love oh crazy stupid love he was like let's watch crazy stupid love
    \item 113. that or if you were going round a corner one wheel would go like that that's what happened to us that's bad and there was one the vitesse one of them at gate duty at school thought this vitesse came up the the roa you know the concrete road approaching the school and thought and just before it got to us the g wh the front wheels went and went
\end{itemize}

Cluster 2
\begin{itemize}  \setlength\itemsep{0em}
 \setlength\itemsep{0em}
    \item 4. cupboard and anyway before christmas we were checking to see how me how much alcohol we had and we've had got we pi we've been given we've bought and it's had duty - free from lots of trips yeah and so said to me aren't you gonna go and buy some then? i said no we've wewe've got we've got we
    \item 14. a quite bad one to actually because the plane got delayed by about four hours when it everyone was already in the gate when it happened the plane got hit by lightning so obviously everybody started tucking into their duty free booze by the ti like within like an hour people were like smoking in the gate they had to send like armed police to like supervise no way and like by the time we got on the plane
    \item 16. well we did u when when we went to lanzarote yeah we did i mean years ago when me and used to go to tenerife we had loads and loads of time to look round duty - free you don't now mm no you've hardly any time have you? no no don't understand why it does that though cos our was in the what? monsoon we were
    \item 19. t you? the thing is you have to keep pulling it out all the time yeah showing it to people you know and the boarding pass and show it to these people show it to those people get duty free you've not go to have too much in the bag have you? cos you can't get at it now i'm bringing this trolley so i've put a few clothes
    \item 21. them drive well exactly no well i'll pick them up ye oh right yes yeah i said i'll pick them up yeah cash in on the yes old yeah tax exempt old yeah well not duty free but yeah get yourself a nice bottle of spirit and mm he smells nice well on twenty - sixteen we'll have to work on getting them you get a passport you can bring him with you yeah
    \item 30. yeah that's fair enough but like you said there's extra bits isn't there? tax and stamp duty and what have you? yeah but they're coming out of i've managed to convince dad to pay the stamp duty out of our savings oh that's nice did you manage to convince him
    \item 31. said there's extra bits isn't there? tax and stamp duty and what have you? yeah but they're coming out of i've managed to convince dad to pay the stamp duty out of our savings oh that's nice did you manage to convince him or did you just raise the subject? i told him how much you told him how much it was twenty - seven thousand and
    \item 36. erm her snapchat today wanted me don't made me want to throw my phone at oh don't don't we are going absolutely batshit about it did you see her duty free purchases as well the other day? in her with her ted bated baker and all no i i literally just saw her tiffher tiffany bag don't because she g oh no she had
    \item 37. at school that's an immediate impression immediate yeah impression on you no matter what the circumstances so what turning up in hot pants and smelling of gin? well and drinking on the plane yeah and getting duty free bottles clanking clanking in the see through bags hello hello erm yeah yeah actually that's a good point on our erm hand luggage we should make sure we've got a a
    \item 38. would have to go in the main luggage cos i'm not carrying it you can't put it in the main luggage no when we get to the airport you can't cos duty free is the other side so you've already checked in your bags no i mean when we arrive in muscat oh there might be some just as you come out? or put the not so
    \item 55. they put extra on the on the petrol? no cos no they stopped it for yonks didn't they? they still kthey still kept it they still kept it yeah the duty the thing is they've taken a penny off tax on beer and spirit and you're thinking well that's nothing is it you know compared with what people drink i mean they're not
    \item 56. what else did the they bring back for you? no no just oh there was something er the shopping bags a nice shopping bag a shopping bag yeah plastic carrier bag they told me they got that from duty free on the way home oh yeah? an emergency present they said yeah oh you had a keyring didn't you? yes did you have a keyring? you know not not a high quality
    \item 68. it would just be nice if with bearing in mind the sort of like eighty percent of the price of fuel is actually tax mm oh well the special thing is you actually pay vat on the fuel duty righ oh okay right um you pay tax on the tax? yes that makes sense ha in someone's world yeah sort of it must be quite disconcerting having that guy sitting on his bike
    \item 74. s great you know he could cyc yeah i mean cycling cos fares in london are expensive mm but anyway they they like it but they'll have to pay seven thousand five hundred pounds stamp duty yeah oh what? it's ever so stressful that is a bit and i sort of think you know i sort of scratched around and i i said give five grand yeah you know towards the deposit
    \item 75. goes bonk straight to the government yeah yeah and it's so annoying yeah it makes you sick doesn't it? yeah i mean oh why they why they do this i i mean stamp duty but i mean that's a terrible tax for yeah people who haven't got much money we'll it is it's just cos like both of them first time buyers but they changed
    \item 76. t having that mm but if you went to america for instance mm or if you went to america and bought a load of stuff and tried to bring it back then you'd have to pay yeah duty on it yeah right i think they allow you something like so it's not that cheap after all? fifty quid is it? yeah no not a lot not a lot cos that's
    \item 77. bay from america yeah you have to pay for that if it's cheap you're alright but if it's anything out you know mm you pay a whacking great lump sum of duty oh and they make you pay it before you get it anyway so they do yeah mm so there's no paying it one once you go out they don't say right now you owe us something
    \item 80. mm erm and then and to think that he's saved enough for a deposit down there which would've been substantial yeah he's got he's got enough for a deposit stamp duty on top of that like all the legal fees on t mm like when they were yeah when they were looking for a house they have a substantial amount the only thing that was waiting for was that extra six
    \item 81. tell you the story it's fine yeah put them back in the bag go on which part? right so like half six this morning erm you know like as you're going through like duty - free and you've got all the perfumes? there's like a new gin and the bloke was there just like setting it up mm obviously so it's half six in
    \item 87. car wasn't a normal car goodness cos it was really small and it had a and it had like a fire but the fox in t on it wasn't on erm speeding ticket duty? no no they were just they were just they were trying to find erm i'm confused the cable was being nibbled by something i'm really confused as to this bit of the film
    \item 88. i'm confused the cable was being nibbled by something i'm really confused as to this bit of the film erm it was at the end and they said and right and fox parking duty and they're like what? and he said only kidding there's this right here's your mission there's this car ah there is there is a person driving the car no it
    \item 99. that? if you buy a second home or potentially if you buy a new house before you've sold your old house or anything like this mm erm you have to pay higher rates of stamp duty oh yeah i've heard about that i'm not i wasn't sure of the erm how it was hedged about with erm advantages and restrictions so cos you know sometimes you
    \item 100. s paperwork is the same and then i try and find out if any of the offenders are here mhm and if they are are they represented by a solicitor er er whether they want to see the duty solicitor which is a free service uh - huh that they can have and er you know if it's a youth have they got a parent with them? er are they known to the youth offending
    \item 114. well you will if you sell your house and no i won't no the limit for is three hundred and sixty thousand ain't it? oh yeah is it? for i for estate er duty inheritance tax yeah yeah but there's inheritance tax as well that's what it is yeah inheritance tax estate duty yeah yeah same difference mm yeah it don't work i'm not like you
    \item 115. thousand ain't it? oh yeah is it? for i for estate er duty inheritance tax yeah yeah but there's inheritance tax as well that's what it is yeah inheritance tax estate duty yeah yeah same difference mm yeah it don't work i'm not like you people well there you go there you go yeah like what? well well we put this up for sale didn't
\end{itemize}

Cluster 3
\begin{itemize}  \setlength\itemsep{0em}
 \setlength\itemsep{0em}
    \item 1. out and all the sheets some of the sheets have holes in them oh man that's really crazy for real erm but the other thing is that like is it's the husband's duty to satisfy all the sexual desires of the of the wife oh fucking hell so so like which is like sound tough man so they have to so at any like at any moment no matter what when they wanna
    \item 2. wants to see him? yeah yeah yeah that's not allowed is kind of funny and cool the sheet thing? we can do it if you want no not the sheet the the husband's duty oh yeah yeah no it's good no it's just funny why it doesn't work the other way round yeah exactly unlike the other rules usually like when they have to shave their head after
    \item 6. i think if they just keep er cos they are legally obliged to provide for a full education or at least access to a full education i think you could be arguing that's breaching their duty of care the school can't legally stop them mm but social services probably could yeah i think it would be a case of if you're doing it and it you know and it's not
    \item 7. go - go dancers wear er brassieres right the law was later amended erm to reflect that it was possibly unnecessary for male go - go dancers to excellent i liked that the erm statutory duty for shit right sort of stated not only which but they're also responsible for the of parking signs city right erm westminster yeah thanks very much i may or may not know have as much fun as
    \item 13. clearly a bit of a communication problem here mm if i have to drive to sheffield mm to collect it i will be claiming back my mileage absolutely i mean ultimately they have a resp um a duty of care yeah you were a student once yeah do you know what i mean? it's not like you you you exit out the door it's not a big school liz really? it
    \item 25. he'd be sick of panto by now yeah his his friend's in it so he wants to see it oh they all go to see him i think so yeah you know he felt duty bound i don't know i quite like he's quite nice i've not heard of he was out in the yeah that's a new one isn't he? right down the
    \item 42. ' t need to be more tired don't need you to be more tired and the money it's going to cost us to raise another yet another one at that point it's his duty to have a vasectomy yeah had you ever considered that? no it sounds horrible yes yeah sounds p it looks like a really sounds painful i don't really wanna get it done's brother
    \item 45. grab skills cards when you need them right actually we're a bit screwed has nobody got red? no i've got red only's got red so anything that involves red is's duty okay which er so if he was a cylon we could be really in quite a lot of trouble already yeah's is piloting so er some of these cards say like you can attack up to
    \item 50.\# \#o they all maybe they all go for eggs yeah but it's er something's got to be done because as one of my colleagues said you are encouraged to shoot rabbits it's your duty well we all know that but the seagulls are protected mm well that doesn't seem to make sense i'm not saying it's the seagulls' fault it's obviously
    \item 57. you put erm sex over y erm over kin yeah but it happens obviously he's i guess he's a bit black and white he's thinks well i've done my duty and yeah he's he's he's a bit hurt but then again he er he he sh you know his his daughters did f erm i can't speak for but and were
    \item 58. and would do and you so you know er he might expect his grown - up children to look after him but they're not going to want to do it no they might feel a sense of duty er or or maybe not i think i think er probably would ju certainly taken him in and she certainly is the one to deal with his his mania on a day to day basis so yeah and you think
    \item 71. can't regulate well they're not gonna they're not just gonna say here's a massive amount of chocolate fill your boots are they? well they might do they have some some duty of care for children i don't know the thing is is it's not their problem is it? thing is the thing is they they can fill them up with shit and then hand them back
    \item 72. ? a bit more prominent as well as in in a song oh yeah i'm not basing it purely on that she's er she's a cool cat mm i sort of feel duty - bound to say rum tum tugger because i know the guy who played him yes but but actually i think my favourite is um the magician cat um mistoffelees mistoffelees he
    \item 79. like the search thing was weird as well the brother and the ex - boyfriend that was honestly there was definitely something suspicious about them the boyfriend the boyfriend like not the boyfriend the brother like you has a duty to find out the truth you need like why was he so defensive of the police? and saying how much he loved them and whatever when even if you think it was avery you'd go maybe he
    \item 95. s going on in and my mum phoned up to make an appointment and they said no he's left we were like oh my god cos i i got hair see now they have a duty to tell people where people have gone i got my hair vouchers yeah so i could get it done in with him cos i get it done so infrequently and then he
    \item 97. that because surely they need to their customers be be no be be because they ha they ha they have to look there's a the way it's put is this and they say their primary duty is is to the person they're caring for the the the the carers feel that they can't look after that person without a risk that she will fall be injured yeah break her hip and
    \item 98. could be the end of her because people of gra of granny's that happens to that happens to easily yeah true yeah um so so th there's that one and the second the other duty of course is to the carers yeah yeah who mustn't hurt their backs and so on but of course they sign a contract saying that they won't do the lifting yeah but the thing
    \item 103. cause i have though you are responsible i've got a ridiculously bad back i just don't know i do get bad backs just tell you're responsible you're you have a duty of care have you heard of the citizens advice bureau? yeah my new best friends yeah you know they're the ones i called about eh the smoking issues is that mine? i didn't realise
    \item 104. what did they do? er basically the penguin classic translator hates the book oh it's rather funny he writes a long sort of translators note at the beginning explaining how he felt it was his moral duty to make up for all of the terrible things in this book which does make you wonder why he chose to translate it in the first place but there you are presumably they told him you're going to
    \item 106. mum there's nothing we can do they're not changing it they just she keeps saying why do they keep feeding her? i said because they have to they have to it's a duty of care yeah i know but if she lived on her own she wouldn't be here and then it'd be so much better for she said oh she'd hate it if she could see
    \item 116. of money you know but they used to send it they used to send cards to each other and never met never seen the people for donkey's yonks mm and yet she felt it her duty to send them cards yeah and i used to have to go out well i posted all your cards didn't i oh yes yeah oh no i didn't did i i left them in the front
\end{itemize}

Cluster 4
\begin{itemize}  \setlength\itemsep{0em}
 \setlength\itemsep{0em}
    \item 12. into racing games? or fifa? or? i'm c or just? all for racing games fifa i'm prpretty bad at but i enjoy i enjoy being bad yeah call of duty i'm bad at again good but i enjoy being bad you'll probably be like a similar sort of level as me's very good at erm like this kind of stuff have you ever
    \item 60. herself but she has like a really active social life in chat rooms and she's old which is amazing it's like this whole thing of like retired men loads of friends and like call of duty and like xbox games and they just chat on their headsets yeah and like xbox games so it's like retired men sat at home all day just like gossiping whilst like blowing each other up do
    \item 61. women like phone each other up yeah oh yeah just having a chat men can't be like yeah alright barry like what's going on? how are the kids doing? but on call of duty you can yeah yeah they're not really allowed to do that are they? have a natter no no you can if it's like through the medium of call of duty and you're
    \item 62. on call of duty you can yeah yeah they're not really allowed to do that are they? have a natter no no you can if it's like through the medium of call of duty and you're like or if you're in the pub yeah it's like pub or golf like golf is ba like just so men can like hang out and have a chat but you
    \item 63. t it? it is actually like how awful yeah but the idea is that is that you're surviving the evil you're not carrying out the evil mm okay i mean there was call of duty where you had to torture someone and that's pretty yeah pretty brutal i always used to find myself like judging parents when they'd come in with like their seven - year - old boy and be
    \item 64. and that's pretty yeah pretty brutal i always used to find myself like judging parents when they'd come in with like their seven - year - old boy and be like buying them call of duty and just think do you know what it's about? like but call of don't let your call of duty is but call seven - year - old play call of duty it just reflects the
    \item 65. like their seven - year - old boy and be like buying them call of duty and just think do you know what it's about? like but call of don't let your call of duty is but call seven - year - old play call of duty it just reflects the kind of the time we're in like mm call of duty originally i think i don i kind of remember playing like
    \item 66. them call of duty and just think do you know what it's about? like but call of don't let your call of duty is but call seven - year - old play call of duty it just reflects the kind of the time we're in like mm call of duty originally i think i don i kind of remember playing like one of the first or second ones and yeah it was violent
    \item 67. but call of don't let your call of duty is but call seven - year - old play call of duty it just reflects the kind of the time we're in like mm call of duty originally i think i don i kind of remember playing like one of the first or second ones and yeah it was violent it was you know but it was no different to any other game at the time yeah
\end{itemize}

Cluster 5
\begin{itemize}  \setlength\itemsep{0em}
 \setlength\itemsep{0em}
    \item 3. The stapler, yeah. So eh, we decided that after, as a form of punishment, students can't be trusted with staplers. Em, we are looking into maybe like a standard sort of heavy-duty stapler. Oh, are we? But the cost of it is a bit much, so we should like chain them 'cause we've got these holes in the desk. We should get like a chain and, yeah.
    \item 9. Alternator off. Erm, what else? Something else? Er, alternator, power steering pump, and then to get to the bottom, where that goes on to the er, there's like a spring, heavy duty, fairly heavy-duty spring. Yeah, two erm fittings where bolts go fully. Yeah, I had to take the whole erm casting off at the engine, bla, all of those off, yeah. Yeah, take.
    \item 10. Erm, what else? Something else? Er, alternator, power steering pump, and then to get to the bottom, where that goes on to the er, there's like a spring, heavy duty, fairly heavy-duty spring. Yeah, two erm fittings where bolts go fully. Yeah, I had to take the whole erm casting off at the engine, bla, all of those off, yeah. Yeah, take the whole casting.
    \item 11. Thing I sort of thinking about 'cause they're so handy, aren't they? Yeah, 'cause I use a, well, this one has got a really thick blade and that's a heavy-duty bandsaw and that, you put different blades on. Yeah, you need to get a new blade. I got a metal bandsaw. Have you? Yeah, for metal. Yeah, yeah. But now, but that, did that lift up?
    \item 15. Is about dropping stuff and, and was saying she'll be available. Yes, but you're not mandated to come. So I mean the ten weeks, it's not a, it's not a heavy-duty ticking the register kind of thing, at least I don't, I didn't get that impression. But er, it's there, you know, for people. But you know our experiences when we have these drop things is.
    \item 17. Know, I think it's called a fruit processor. What's the difference? This is a blender, it blends. Yeah, whereas processors, it like chops it differently. Oh, so these are more heavy-duty than these. They'd make a smoothie and something else in this. Very cool. What else are we going to put in this? Put a bit of cumin? Do you like cumin? Yeah.
    \item 26. Now it looks like a jumpsuit. It's not 'cause it's shirt and c, like combat-type trousers that have got like cuts there for the pockets. Mm, big black heavy-duty belt, um, navy shirt with the square here that says the, mm, like that has the ring on it to say if you're an officer or not. They're all navy, then this bit on the.
    \item 43. There. Mm, we did put a lot away this time. Took us a long time, but cleaned up all the rest this morning. They were soaking wet, were they? Absolutely drenched they were. It's a heavy-duty outside. I don't know, 'cause you didn't get up early enough to see it. I didn't get up early enough to see it, no. Went to bed late. Went to bed a.
    \item 51. I don't know. Re, I don't know. Uh, which one do you prefer? Come here, yeah. No, it's lovely, isn't it? And actually it's, it's heavier duty so it would be better, as you say, if someone's buying the glass tut containers. I like both of them but er, I think it will be easier to stamp the logo on the cotton, yeah, definitely.
    \item 52. Put the ribbon through the, 'cause that's a scrub in there, obviously a lot heavier 'cause it's gonna be full of products, see? Yeah. No, it does need something more heavy-duty, doesn't it? Huh. Hm. And the scrub's gonna be heavier, isn't it? 'Cause this is empty. It's gonna be f, yeah, really heavy because it's full of salts and.
    \item 53. It's full of salts and, and it's in a box. They come in a box as well. Well in that case, you're definitely gonna want a bag that's more heavy-duty so that, especially, well, if it's in a ba, if it's in a box that'll probably protect it, in fact, quite well. But I know this isn't full and it's.
    \item 59. You asked us for a long thing, that's right. So I, so I bought this socket, which is about that long. Yeah, and then I had an extension that long, mm, yeah, and then this heavy-duty ratchet on the bottom. And I tell you, that was as much as I could do to hold that, the weight of this, the whole lot, spanner stuff, yeah, let alone turn it. Yeah, and then you think.
    \item 78. Said, or I'll get me all the gear. He sees I got a green card, a green ticket, and he says, you're no good here. This is all, this is a heavy, heavy grade, really heavy-duty job, don't they? Mm, and like these idiots sent me down there and you have to go. You, no, you have to go. They don't change, do they? No, these dole places, you.
    \item 101. Off, it's illegal to have open anyway, isn't it? It's not a fire door. It's not a fire door. If it was a fire door, it should be a heavy-duty door and that's just wood, not window, not with glass in it. Yeah, well, the hinge I think might have made it heavy duty. Oh, did it? It's a solid door, solid, but the.
    \item 102. Door. If it was a fire door, it should be a heavy-duty door and that's just wood, not window, not with glass in it. Yeah, well, the hinge I think might have made it heavy duty. Oh, did it? It's a solid door, solid, but the hinge always uses the catch and it was so, it made it heavy 'cause it was a heavy weight. No, it's def, definitely.
    \item 107. T gonna be any good. It ain't gonna be any good, no. Or it's gonna permanently annoy you 'cause the only time you want it is when you've got some heavy-duty stuff to do and that ain't gonna be enough. Yeah, yeah, that's right. Yeah, well, how thick is the, the strimming, er? The string on it? The stri, yeah, erm.
    \item 111. In my, under my kitchen sink, so we ended up picking them up from whilst we were at Asda and then it was like, oh, do you want them fleece-lined? Do you want heavy-duty? Who knew there was such, who knew there was such a choice? Do you want plain pink? Or yellow pink with a lining? So he, he's ended up taking, oh, I think it was.
\end{itemize}

\subsection{All contexts for \textsc{planet}}

\begin{itemize}  \setlength\itemsep{0em}
    \item 1. cos that's not a real thing a hoverboard it is a real thing what about a holiday home in spain? w yeah or we could build a giant mansion around the entire planet i don't think we've got enough money to do that mm we would enter the lottery no we would be in the lottery group and find out exactly which mon and exactly what money we need
    \item 2. they'd all chosen pink shirts yeah so there was er how funny and all sat there in pink shirts huh and then er walked in pink pink pinstripe must've been on the same planet then why did i just choose juice for then when i've got wine down here? that could've gone wrong huh gone wrong in the head your base looks alright actually mm it looks like the
    \item 3. ' t understand like at his age why he doesn't think that that's a problem? like having gammy teeth for the rest of your like sixty years that you're on the planet not even gammy teeth how can he possibly think that you'd be interested at all? even a little tiny bit i don't know it's a worry it w it worries me if
    \item 4. is there i think that's the problem he gives all the big i am but i don't really think er he's really on the same planet no i don't think he's got any brains that's the issue here mm oh my mystery meat's stuck come on mystery meat mystery meat oh huh oh huh i only wanted two
    \item 6. on whatsapp erm or viber or whatever uhu free free messaging erm so yeah i'm messaging back what the very fuck? what what are you on? uhu what planet are you on? what are you on about? when did this become about money? this is about you getting to choose the bouquet yeah that you would like to your specifications nothing to do with money of
    \item 9. yeah we'll see we'll see i just want to go and have some adventures i don't blame you you only live once don't you? you're only on this planet for a certain amount of time just go an experience it and enjoy the time mm hm perhaps have memories if you live long enough if you don't at least you enjoyed yourself that's the idea
    \item 12. about five years or maybe longer mm but they've travelled a lot hope to see you this year ah even better than flying will definitely be a cruise cruise i don't know the cost what planet does this guy live on? i know the two but two of you would be ideal for but is that better? four berth berth a four - berth cabin with bathroom toilet and shower and friendly friday?
    \item 13. blimey yes er i'm st i'm staggered there isn't a cure for c cancer yet mm and of the course the the the there are just to many people on this planet now and because we don't have the world wars that we used to do where there was a natural selection process as it were culling yeah there was yeah men and women mostly men sadly but er
    \item 14. yeah i don't know about that stuff but there's gotta be life out there because we are there's life here yep end of i agree you know so there has to be another planet ninety - two million miles from its sun that can sustain life i agree i mean if you think about the fact that you know the whole gestation of a child mm and the creation yes thing that '
    \item 15. whole gestation of a child mm and the creation yes thing that's miraculous isn't it so you're thinking you're telling me that yeah life can't exist on another planet but that can happen? yeah it just doesn't doesn't what do you mean? it isn't really supposed to make sense it does make sense what? aliens coming to the planet like why were they there? doing weird shit and they're not aliens they are yes they are no they're not no they leave in a spaceship no they don't they do they no
    \item 17. it's really fun did the old biddies take to you being there? or were they a bit? they loved it because w okay and i were giggling away to ourselves er absolutely on another planet and they kept looking over and smiling at us like okay as if we were special as in like oh right those girls bless them are special yeah yeah i think cos i was always going the wrong way
    \item 21. mm have to find out no don't wanna and the two people that er she brought with her three people that she brought with her when she first came they just disappeared did they? off the planet they have oh can't find them at all really? can't find them at all mm anyway how's your woman? how's my woman? oh she's alright yeah?
    \item 25. s just a bit scatty yeah which is the same as you know money isn't an an issue she's not tight she just isn't engaged on the p on the same planet yeah how much money would you wanna have if for example you did wanna go travelling? say it's australia i would need to have so on entry into the country you need to have five thousand dollars
    \item 26. it and it's sort of ironic isn't it? that kind of the likes of you who have been mauled over by major surgery but we're glad to

 be here on the planet really never mind what we look like you know we're glad to exist erm yeah so it's a it's a strange er you know it's a very long journey you go
    \item 27. okay erm yeah i mean i think that is a huge issue and i suppose i think a lot of people feel they've just got to be so grateful that they're here on the planet yeah at all yeah that all those other things just kind of you know they can't whine about those things because they're just quite glad that they're here at all yeah on the
    \item 28. at all yeah that all those other things just kind of you know they can't whine about those things because they're just quite glad that they're here at all yeah on the planet and so unless but i kind of do somewhat feel that erm but then on the other hand you see but but but i my husband's younger than me he's only in his forties
    \item 35. for someone else i become the most assertive person in the world yeah yeah and my problem is i don't think i expect anything of anyone i think i expect less of anybody on this yeah planet than anyone on this planet i've ever met yeah in my entire life yeah but one thing i do expect it never happens but i do expect i expect someone to stick up for me mm and mm
    \item 40. i just find it sickening and then yoda throws the bowling ball down the er the bowling alle the lane and the woman goes oh how did he do that? you know like w what planet's she been living on? yeah how did she know yeah she's like looking at little i mean everyone knows yoda yeah looking at yoda yeah as if he's a normal person
    \item 41. kind of understood that really her situation isn't that bad yeah d'you know what i mean? absolutely and she just wasn't getting it not at all yeah she lives on a different planet though doesn't she? i mean it is yeah she
\end{itemize}

Cluster 2

```latex
\begin{itemize}  \setlength\itemsep{0em}
    \item 5. to uni it's mickey mouse ah thank you i love it i i don't know why the box is so big though the box like oh it's huge they're this planet with their cardboard boxes yeah's never gonna work in gift wrapping tell you what that that'll do there you are when when i when i was wrapping's present i forgot how to wrap co
    \item 8. ah yeah proper ridiculous it was mars weren't it? hasn't quite worked out that's mars well i don't think it is even mars i think it's just another planet that's uninhabitable if you go out of the bubble of oxygen yeah hilarious and at the end they are about to explode they're like coming out their face it's so
    \item 16. go what the hell do we want with these people they're like what stay away from them they'll just er kill themselves off in a few minutes we'll just keep checking on the planet and we can move in when they when they have destroyed us but there is there's some great theories the er conspiracy theories out there and there's gotta be t s you know there is some
    \item 24. it seems pretty unlikely i mean the that's there's er time for that yeah when i'm old and decrepit if we haven't destroyed the planet by then erm planet yeah er yes so i don't know i think i'm just gonna save like a beast like with these recordings it's obviously been really great to pay off my overdraft and i
    \item 34. the geography of light the geography of light? i'm okay for the geography of light go on okay the geography of light light light light light how many times further away is the sun from our planet compared to the distance the moon is from us? two hundred four hundred or six hundred times? four hundred? four hundred right do you wanna play again but without the chips? can do yeah cool yeah
    \item 37. jupiter? i don't know pluto he's like from pluto yeah he's from pluto he he's from the comet and he got really pissed off that they're a dwarf planet now but oh no and yeah culture is obviously has a huge impact on it yeah and but that's not to say that culture isn't real the fact that culture has an impact on people er
    \item 39. was like fish swimming and it felt like you were swimming with the fish yeah that was good but then when you so do you remember that one we went to in the space centre? the oh the planet one? where you like lie back? yeah yeah that was good yeah i dunno what they call that? like those ones are good but then when you just go and see like a film on the
    \item 54. in a spaceship no they don't they do they no they don't yes they do er what are they? trans tran transylvania wait no they're trans sexual they're from the planet transsexual in the galaxy transylvania or something or is it? or the other way round no they are transsexuals and they're from transylvania no no no transsexual
    \item 55. s belt mm mm no i don't know er anything really that's why i wanted to go along yeah ask them how many of those stars could be a sun they all are to another planet aren't they all? they're all s s stars yeah but how likely is it that there's another planet like this off one of them s stars oh actually the the odds without actually
    \item 56. how many of those stars could be a sun they all are to another planet aren't they all? they're all s s stars yeah but how likely is it that there's another planet like this off one of them s stars oh actually the the odds without actually knowing the odds must be very much in the favour mm mm cos there's just so many of them it's
    \item 61. just wanted us to remember but why? i don't know when i was little we also learnt pluto but they've got rid of that now they said it was too small to be a planet pluto? i was alive when it was it was when you were born pluto? it was the thought of as a planet for a while when i was little because i when when i was about maybe three
    \item 62. ve got rid of that now they said it was too small to be a planet pluto? i was alive when it was it was when you were born pluto? it was the thought of as a planet for a while when i was little because i when when i was about maybe three years old i had a big erm solar system thing oh you did you did have a big chart about it yeah yes
    \item 63. maybe three years old i had a big erm solar system thing oh you did you did have a big chart about it yeah yes and there was pluto on it mm although now we've discovered planet nine i think well we haven't but there are certain signs of it mm either that or's making it up oh i don't know in our solar system? in our solar system a
    \item 64. think well we haven't but there are certain signs of it mm either that or's making it up oh i don't know in our solar system? in our solar system a ninth planet or tenth planet including pluto which i think that seems a bit mean doesn't it to? yeah just take taking a planet i don't know how they decided out of the solar system you know
    \item 65. haven't but there are certain signs of it mm either that or's making it up oh i don't know in our solar system? in our solar system a ninth planet or tenth planet including pluto which i think that seems a bit mean doesn't it to? yeah just take taking a planet i don't know how they decided out of the solar system you know that it was
    \item 66. ' t know in our solar system? in our solar system a ninth planet or tenth planet including pluto which i think that seems a bit mean doesn't it to? yeah just take taking a planet i don't know how they decided out of the solar system you know that it was too small and it couldn't be a planet but anyway yes so that chart was wrong and all the books
    \item 67. doesn't it to? yeah just take taking a planet i don't know how they decided out of the solar system you know that it was too small and it couldn't be a planet but anyway yes so that chart was wrong and all the books that they'd written about planets were wrong so do you know what erm a pin - sized thingy of a blue planet of a
    \item 68. a planet but anyway yes so that chart was wrong and all the books that they'd written about planets were wrong so do you know what erm a pin - sized thingy of a blue planet of a blue planet erm was the same a pin - sized thingy had yeah i've told you this the empire state thing yeah of the pin - s head - sized blob of
    \item 69. yes so that chart was wrong and all the books that they'd written about planets

 were wrong so do you know what erm a pin - sized thingy of a blue planet of a blue planet erm was the same a pin - sized thingy had yeah i've told you this the empire state thing yeah of the pin - s head - sized blob of a? oh what
    \item 70. matter of erm matter of a star weighs erm the same as three empire states three empire state buildings so pretty heavy yeah yeah but not actually of a star of red or blue yeah a huge planet which is a star but it's n it's what? it's much bigger than the sun and much more powerful it's blue which means it's extremely hot so that '
    \item 71. mm although there is i can't bel although mm? in my it can't be true book oh yes is wrong cos it i'm not sure if it has pluto as a planet i don't think it does but it definitely doesn't have planet nine i know because it's a new discovery my love yeah this is my rice and i would like you to try a
    \item 72. can't be true book oh yes is wrong cos it i'm not sure if it has pluto as a planet i don't think it does but it definitely doesn't have planet nine i know because it's a new discovery my love yeah this is my rice and i would like you to try a little bit of it please both of you do you know the song that?
    \item 73. and what's the actual ultimate object of the game? to beat handsome jack and have you met him yet? kind of who is handsome jack? handsome jack is er well there's a planet and erm there's a vault on this planet that has been discovered and it holds a very valuable alien resource called eridium and everyone in space wants eridium and this planet called pandora has a
    \item 83. very good book oh okay right then i like people i like people more i can redeem myself from that two speed is of the essence what what am i on? ah you can have these the planet? world i can't do you want to do the timer? yeah hang on i'm going to get the recorder out cos i what are these ones? they're done okay concentrate
    \item 85. \#\#wp oh it is gawp what's a gawk? or it can be a gawk sounds like a some alien from er it can be gawk or gawp the planet zog it can be gawk or gawp depending on er how strongly you're doing it i reckon it's a weird word isn't it yeah oh yeah it is gaw
    \item 87. like as if it's filming it er twenty - five frames a second or something you can then have twenty - five photos yeah per second but i think that on on to choose from on frozen planet it was because like say if they were waiting for like a um tut a whale to like burst through the surface yeah you like you you can't guarantee that it clicks on time no by the
    \item 98. does make you think yeah yeah course gold is the only metal but they mine it out the ground don't they? yeah but that's only cos it's come together on the planet you see see planets are different from stars yeah they're formed by there's all this dust flying around in nebula right where you hear about like the horse head nebula that's right yeah they
    \item 99. get it coalescing right? yeah and as it coalesces it build up its centre of gravity more and more which means it drags other stuff into it and and eventually it becomes a planet you know with just made out of dust in space and the bigger it gets the more it starts pulling it in yeah no they they were i mean that's i mean you know i'm probably
    \item 103. ' t really do any they you know it wasn't a battle it was just a traditional alien thing it was just a really weird mind thing ha yes and that bit with the er the god planet was weird the god? oh the god oh yeah oh and then they all came back to the yeah ship oh that was weird that was fucking weird mm not in a particularly nice way just in a slightly
    \item 110. can smell it but like i dunno what's natural about it but it just stinks who is the greatest empire? what every like be like if quantum mechanics governed reality? why is a planet? oh oh my god i forgot i was drinking this fanta and i nearly put a snapchat oh my god everyone knows about this fanta oh wow's is interesting oh yes and kind
    \item 113. on all the channels huh look he's bringing the flower back to life aw huh that's better that's why i think if there were aliens they wouldn't be bothered with this planet they'd be i'm sure mm an advanced civilization oh that's you i wondered what it was ghosts? mm ghosts in the machine huh god you wouldn't s ever go back to
    \item 120. northern france associated with the second world war the beaches dunkirk yeah er it's a like a west indies they and a place in the west indies the a it's a it's a planet with a rude name uranus yeah it's a er a blue colour on monopoly very posh place angel islington no they are dark blue a very famous place very expensive mayfair yeah er it '
\end{itemize}

Cluster 3
\begin{itemize}  \setlength\itemsep{0em}
    \item 7. \#\#cksack well no because take rucksack to sainsbury's do you? yeah yeah and walk out with them full yeah tut tut yeah cos i'm saving the planet man i've never been stopped no? but then i do have i'll tell you what if somebody ever stopped me when i wasn't stealing something i would be fucking pissed off probably because
    \item 19. ' ve got national geographic oh got signed up to what's this you're on? all these were it's the same thing they're all just different magazines wow pulse on the planet i bet two minute sound portraits of planet earth nature and science so before we were just saying yeah go and t talk to practitioners speaking er yeah listening sorry they've got oh that's good they
    \item 20. up to what's this you're on? all these were it's the same thing they're all just different magazines wow pulse on the planet i bet two minute sound portraits of planet earth nature and science so before we were just saying yeah go and t talk to practitioners speaking er yeah listening sorry they've got oh that's good they've put a lot more on here
    \item 23. boyfriend? it seems unlikely it seems pretty unlikely i mean the that's there's er time for that yeah when i'm old and decrepit if we haven't destroyed the planet by then erm planet yeah er yes so i don't know i think i'm just gonna save like a beast like with these recordings it's obviously been really great to pay off my
    \item 31. they're like little mini seahorses weren't they? did you see that episode of south park where er they buy a bunch of sea monkeys? yeah and they like welcome them to planet earth and stuff and then they actually do evolve and like they go through the stages of civilisation and like make pyramids and stuff it's really funny i'm sure we've had this
    \item 32. to pay me for going underneath it mm and then she said her husband was coming back flying back from new zealand and that was the end of the discussion cos i said what was that doing for planet? mm how much money does the government put into research in comparison to the subsidies that are given to these erm wind turbines and things? well a pretty good question i i don't honestly know
    \item 38. re so successful we developed such large brains and now we're and now we are too complicated to compare to primates who who yeah yeah who are much more simple have you seen rise of the planet of the apes? caesar yeah there are ways to look at primate anthropology in the right way and there are ways to do it the wrong way and most people do it the wrong way in that they get
    \item 42. drama with john w john wayne wow there's nothing on what is going on? what's on the bbc? on the beeb? oh god formula one ooh no abort human planet formula one mm don't want john wayne what's a night to remember? titanic oh god pooh's heffalump movie pooh's heff yes good good that '
    \item 45. was like quite quirky um give you a handful uh hmm um as opposed to just the one that you need please give them the eagle eye evil eye oh um we're trying to save the planet or something like it was a really cute little thing and i was like no way that's please please give our workers the evil eye so if they basically thrust them at you fine will do hence take
    \item 49. that's true as you know yeah you could take your tea with you your thyme tea yeah that's okay i don't really need it but what and they've got animal planet on tv oh nice this channel that i really like yeah no it's not a bad option but the yeah he was saying this guy was quite financially savvy and he was just saying it '
    \item 59. ? so accelerated a lot god four and a half million years ago the earth's temperature exceeded the melting point of iron which is why all the iron turned into liquid flowed to the centre of the planet heated into lighter elements such as and then formed a sort of mantle crust i know it's a very hot volatile place isn't it now? i think we should get erm assessed assessed?
    \item 76. a whale of a time look at that hump what are you gonna do with all that hump? all that hump inside that that hump inside that what are you watching? the blue planet i've got it on dvd i've got it on netflix awesome do you want a gold star? how about i give you a punch in your vagina? how about i punch you in
    \item 77. like native america you like wear all its stuff cos then you like embody the powers of the animal which calling upon the power of ten tigers what? er it's from a cartoon captain planet he calls upon the power of ten tigers i do that constantly and nothing happens i call up the power of ten do you know one thing i'd love in life? would be an animal whistle if
    \item 78. re the evolved western world when in actual fact we're less humane than anybody else mm though the issue i have is this idea of natural selection a certain number of people have to die because the planet can only sustain a certain number of people yeah and obviously because of their they they don't have access to condoms men don men rape women yeah so they're constantly having babies there's no
    \item 79. it's gonna put the all sweatshops it's gonna de out of devalue a lot of stuff yeah when they get quick and cheap and it's got to fuck the planet up you can just you know everyone's gonna have a factory in their own house mm just get a blueprint yeah don't need to buy anything any more just need the yeah the plastic whatever
    \item 80. ? yeah it's like well why don't you just download one dad? that's the that's what i was saying to that's why it's gonna fuck the planet up cos every time somebody erm you know loses something or can't find it yeah oh i'll just get a another one yeah maybe you could er you could upload your car key
    \item 84. \#\#dicted each other and and you know mm right and quite a lot of it was well actually we don't we're not quite sure about this yet right but the general consensus was the planet would be better if more people ate more vegetables and fewer less meat yeah yeah yeah and most people's health would be better if they ate less meat than they do yeah in in developed countries it '
    \item 86. ' ll give you give you a choice of the the best ones yeah cos they do that on like wildlife photography don't they because like you know i saw it on yeah was it frozen planet? yeah to m what was the other one called? to make sure that you've captured like cos yeah yeah cos yeah to m to make it like as if it's filming it
    \item 89. money mm yeah and they get really quite envious of that so they try to get money from them we understand we understand but th half of that problem is because people will just follow things like lonely planet and they'll go to that one place whereas you could go to mm these numerous places but no one fucking does yeah cos they wanna stay with all the other fucking tourists yeah mm the problem with
    \item 106. stuff like that like it doesn't faze her she's just not bothered about it which is fine but for people who are a bit worried about you know either saving electricity or saving the planet saving the planet you know just i think that sort of passes her by and it's hard because she's so scattered like with her mum last wednesday her mum was like have you got the tickets
    \item 107. like it doesn't faze her she's just not bothered about it which is fine but for people who are a bit worried about you know either saving electricity or saving the planet saving the planet you know just i think that sort of passes her by and it's hard because she's so scattered like with her mum last wednesday her mum was like have you got the tickets now are you
    \item 118. of the people there had so that probably i would have a chance when when a a job came up and a job did come up and that was to er produce programmes on a series called the living planet where they were record w er were looking for people er to join a team for a three year project multimillion pounds with david attenborough oh right erm and i thought no i won '
    \item 119. they'll hopefully we we've got another supporter and i say to to th as well would you be prepared to give an hour of your time a week to something if you care about the planet and the people on it an hour of your time just licking envelopes or or putting out some flyers? er but you know you could do something because you meet people and broaden your horizons learn a
    \item 123. it's the new head of greenpeace and he's like greenpeace environmental erm our campaigns the worst mistake ever to have been made is that it's about saving the planet it's not about saving the planet when we get extinct when we become extinct the planet's gonna live on it's just gonna rerereresuscitate and revive and yeah it '
    \item 124. \#\#peace and he's like greenpeace environmental erm our campaigns the worst mistake ever to have been made is that it's about saving the planet it's not about saving the planet when we get extinct when we become extinct the planet's gonna live on it's just gonna rerereresuscitate and revive and yeah it'll live on cos it's
    \item 125. environmental erm our campaigns the worst mistake ever to have been made is that it's about saving the planet it's not about saving the planet when we get extinct when we become extinct the planet's gonna live on it's just gonna rerereresuscitate and revive and yeah it'll live on cos it's a powerful thing but it's about not about
    \item 126. live on it's just gonna rerereresuscitate and revive and yeah it'll live on cos it's a powerful thing but it's about not about saving the planet it's about helping people live in harmony with the earth for for as long as possible without basically making it an inininhabitable place and that was interesting because lots of people when you see
    \item 127. yeah and that one's that's dog shit this one's bullshit the funny thing about crystals is how like alive they are with like the hippy scene which was always very you know planet conscious and about eating wholefoods and it's like and then you've got your crystal collection which you need because of vague erm spiritual reasons but they are terrible for the environment because the
    \item 128. they're very very pretty yeah but don't tell me they have mystical properties i know and especially when the whole rest of the ethos of your life is about sustainable living and saving the planet but you uhu keep buying crystals which is probably well not the like biggest environmental oh my fav my my favourite one problem but is that it's impossible to destroy a diamond it's impossible
\end{itemize}

Cluster 4
Here is the list with bullet points, maintaining the numbering at the beginning of each item:

```latex
\begin{itemize}  \setlength\itemsep{0em}
    \item 0. stuff is that er it's impressive because it's so yeah yeah yeah hard to play yeah and it's so ins cos the drums and the guitar that very few people on the planet can do that yeah and that's the main reason it's popular do you think there's a certain amount of also like rebellion in it? yeah that's true and oh yeah all
    \item 10. ' t he? yes he did which is yeah quite impressive yeah yes and sired a few few children on the way didn't well well they reckon er one in every two hundred men on the planet can trace their ancestry back to g er genghis khan that's insane mm er surely not i mean if you think about it obviously it's a lot more difficult for women but for a
    \item 11. ' ve read in linguistics books they're they are they argue that erm our language is one of our most defining features of what makes us human what separates us from all other creatures on this planet is our ability to communicate and what has well not communication and and language as separate er yeah animals can communicate yeah but they don't have language they don't have the s the so they they
    \item 18. ' s a struggle that one cos i y cos that one's just such a struggle for me cos i was like i was like there's like seven billion people on the planet how can you think we're not having an effect? i just don't understand like that view i'm like how you can think it? mm and also it it it shows that they
    \item 22. stuff you know? i like to think that if humans abo abided not abode abided by the laws of christian agape aboded yeah yeah abided yeah every person on the planet since the dawn of human history i don't think we'd have paedophiles no but paedophiles there's there's something intre intre or people like that
    \item 29. all my fun you can listen to this for a minute he just controls things and now we're blatantly not going to like it even if it is the funniest thing on the planet we're going to yeah literally we're just we're going to be dickheads relent on you my whore are they talking about henderson? ah a prostitute she dynasty she dynasty that '
    \item 30. ve died it's it's miniscule absolutely miniscule well that but look at the money they've made they've brought in injections to every single person on this planet now oh yeah absolutely yep at birth or they you know flu vaccines and they try and put yeah they do pregnant women don't they? it's you're talking multi - billion probably multi
    \item 33. mm well they are threatening to renege on the their share of the debt which is an empty threat cos if they renege on that i can't think of too many people on the planet who are gonna lend will wanna lend them a few bob yeah so it's all gonna happen in two weeks two weeks'time innit? two weeks'time two weeks tomorrow mm that's
    \item 36. the most assertive person in the world yeah yeah and my problem is i don't think i expect anything of anyone i think i expect less of anybody on this yeah planet than anyone on this planet i've ever met yeah in my entire life yeah but one thing i do expect it never happens but i do expect i expect someone to stick up for me mm and mm it's not a
    \item 44. out ways of how to charge people for things how they can make money out of everything charge for every single thing and that erm and that's not right when we're born onto this planet we're born with rights that er you know it's it's your world it's no one owns it but the capitalists own it now so and now the the capitalists are
    \item 46. come back i'm sort of seeing this norwegian guy all quite casual he came up to cambridge like five times i went to norway and then when i came back he disappeared off the face of the planet so that was a start of the where men are crazy yeah he was the one who was referring to me as a girlfriend when i was i went up to norway to break up with him and then realised
    \item 48. thing that people can talk to and pray to by putting their hands together and somehow talking to him wherever he is and and he answers their prayers even though there's about seven billion people on the planet probably about two of those probably at least one billion are praying at that time i don't know not one billion praying at that very time but a lot so yes i don't think he is
    \item 50. whatever's left the little bit of powder that's left stick in a firework we'll get norris mcwhirter and see is this the most used larynx on the planet? start you can do research into larynx i don't care i don't make sure you make that joke at my funeral with a smile on your face and your bright bright suit on
    \item 58. mean yeah because they didn't want to be yeah is the problem it wasn't a choice was it? no and they were an established yeah one of the oldest civilisations on the planet mm i mean we met the dalai lama mm how amazing is that? didn't shake his hand or anything but we had a teaching from him mm we were only for the tea and biscuits it was
    \item 81. to most of the popula and i i actually listen to god anything by him but oh dear don't you've got a clea like cl like aggressively clean - living person on the planet aggressively what? yeah but he's cliff richard most aggressively what? clean - living person on the planet oh yeah but he's still like put botox in his face he has that's
    \item 82. t you've got a clea like cl like aggressively clean - living person on the planet aggressively what? yeah but he's cliff richard most aggressively what? clean - living person on the planet oh yeah but he's still like put botox in his face he has that's not particularly clean hmm well maybe it is i dunno it's very clean botox uh probably kills
    \item 91. somebody could not think that yeah do you know what i mean? yeah how can't why can't you see that that's a problem? er there's no way on this planet that that didn't enter into your head to do and you chose not to do it as opposed to just didn't think yeah it came in your head you thought it wasn't important enough
    \item 94. for bout six months actually and i hated london really? with a passion think it's just dirty em i think it's em it's it's the loneliest place on the planet where so many people seem to be around yeah yeah em eh the the whole element of eh you you're never out of your work clothes for for no other reason then you you may as well stay
    \item 95. never had anybody give me that information before really? i was gobsmacked i didn't think anybody alive now knew that bloody hell yeah i didn't think anybody alive now on the planet would have memories of air raids and going to the bomb shelter and rationings mm and mm mm like cos i just i never ever learnt about it never wow not once that's mad so
    \item 97. it says erm dark chocolate is loaded with nutrients that can positively affect your health made from the seeds of the cocoa tree it's one of the best sources of antioxidants in the planet studies show that dark chocolate not the sugar crap excuse me can improve health and lower the risk of heart disease that's what he put i didn't say that dark chocolate is very nutritious
    \item 105. you further on you know no you know liam spent you know over a year yeah no entirely lots of erm graduate improvement days which i can still think are the worst thing on the face of the planet yeah yeah yeah but how can you work cooperatively in a team of people you are competing against? it's a situation it's a skill which is not acquired in yes because you're
    \item 108. you'll live the best lifestyle you can imagine yeah yeah can't prove a thing can't find anything more about him no he's disappeared he's off the face of the planet but you know what i mean though like in terms of the films that i've seen i mean i watch a lot of films and what you do is that you find the you're you '
    \item 111. do you want to pass that enrichment onto another generation? well what about if we have different philosophies what if i don't feel if i feel a genuine connection to everyone on this planet what if i thought like oh lovely thank you thank you my legacy might not necessarily have to be imparted through my own biology no that i no feel more of a kinship with with the people around
    \item 112. now won't they? at twelve? they're all just coming out all these human beings everywhere you sound like an alien who wants to wipe all the humans off the pl face of the planet i probably am you know? yeah maybe you don't have blood and all yeah i know maybe you've got something else yeah i'm an alien
    \item 121. don't don't worry it's fine sweetie no cos you can't be comfortable though we can't hear them blue whales are one of the loudest animals on the planet communicating they're having a party down there communicating with each other using a series of low frequency pulses groans and moans it is thought that in er good conditions blue whales can hear over distances of up to
    \item 122. it and mm i mean like whaling happens in norway and they're like so civilised yeah yeah and they have that yeah like they're so that's like the the country on this planet that's so broad thinking i read this interview with the head of new head of greenpeace oh no well the head of greenpeace is stepping down and he's being replaced oh yeah
\end{itemize}

\subsection{All contexts for \textsc{planet}}

Cluster 1
\begin{itemize}  \setlength\itemsep{0em}
    \item 3. her because she's not been eating oh no she's not been taking her tablets that she should take she's she's had a bit of a rough time of late within her marriage erm i don't know anything it's just a couple of things has said erm but she she shshe spoke about it because she's so concerned about her and she said
    \item 6. and you build a picture up that you think you couldn't ever n ever imagine imagine she went you're absolutely right she said erm there's a lot to it she said their marriage had been over for years she said erm she said i'm devastated she said because i love she said but won't let us keep in touch with her we've been told can '
    \item 7. so who she's married to yeah yes yeah erm she said we're not allowed yes cos they always seemed to get on very well she said i love she said and and the marriage was over a long long time ago she said and you're absolutely right with your thoughts she said because erm it's my washing machine oh erm you're absolutely right with your thoughts
    \item 16. had divorced him yes she had divorced him then why did that fall apart so quickly? er me really quick me fair me that's the reason why so you went from boring to breaking apart her marriage in the matter of about six months? er longer than that there was a what happened was if you it didn't really kick eighty - three was when i met her i think she might have started
    \item 19. like all those years family laughs and stuff you remember it all he's just no yeah course you would course you would bothers me i think what happens is when when it divides like when a marriage breaks and it divides you you pick up new roots with a new part of a family don't you? yeah yeah like you have with's family he has with what's her name?
    \item 20. the weaker path and choosing not to have children in the first place so i mean look you know at the moment you you and you and in the beginning somebody new you know huh you know broken up marriage and two kids can react we know that we've been there ourselves but you know if wasn't the sort to make some effort mm towards and mm because he loves you mm because when he takes
    \item 23. prprobably erm erm your become an alcoholic on the local banker's draft or whatever well they obviously the the breakdown of your er er your mum's and i's marriage forged you to be the kind of person you are now you would've been different er like is different before this w was very extrovert mm you were not you were you were but he was
    \item 32. they before they come out of the green room shall we say how many questions ddo the does the p celebrity or whatever are allowed to sort of say i don't want to talk about my marriage don't want to talk about my divorce and you're not to mention my mother or whatever it might be don't they? yeah well there probably is quite a bit of that because don
    \item 38. what people were telling her um yeah so i think for those reasons you know that was obviously a massive part of why she moved out of so yeah um and ultimately what led to the end of their marriage but yeah mm i think because she had a really crappy upbringing mm a really really rubbish time and um it's amazing like you know told you about that book they fuck you up like it '
    \item 41. ' s very different oh right a lot of them don't move around they s live with their family yes and husband and they can but that's good but it can also break up the marriage can't it? yes depends yes that's right cos pilots'wives have always had to do that haven't they? they couldn't move with their husband right but then those
    \item 42. that? i mean i knew it was yeah but i just i just couldn't mm quite get it i i grew yeah i grew a moustache first did you? this was after my marriage had broken up so oh er didn't er i never knew you yeah you don't with a beard and that's it well she knew me because we as as no but as as a
    \item 48. having a discussion and then yeah i didn't and a little bit of an argument as well mm erm and and i were within that twenty oh right and er there was a i then my marriage ended and was coming to lon he was coming to the uk for something i think he was coming to london to see a friend so he was still in dublin at this point? mm yeah mm and
    \item 52. least she gives that other side of it but with and erm again they've erm they've kind of stood back a bit mm and er cos he was in a very unhappy marriage i think for a long time erm but he wanted to do the right thing you know? yeah yeah kind of comes out more doesn't it now i didn't realise it was that '
    \item 53. ' m sure she's alright i'm sure she's fine but and she might well be but bec oh but went through quite a sort of he's gone through a tough marriage erm and he he has he wa he was really depended on wasn't he? erm depended on him entirely yeah and wanted more and more and more yeah each job more money you know?
    \item 57. kids myself mm hm in a way it was a benefit in one way because some of them used to go home and i'm not saying their marriages were on the rocks but to have their strong marriage you're got to put a lot in to it yeah yeah and so that means you've got to be eh either understanding or say you have eh not in the right frame of mind one way
    \item 83. kids yeah yeah it's not very nice yeah a bit bizarre mm and then i suppose if you just segregate the sexes then you can't really expect people to have like a normal marriage afterwards you know? that's true cos they don't know how to relate to each other oh cos they're segregated you mean just with social situations and yeah yeah so then the
    \item 96. maybe not no er not single sex single parents what those chi what those children turn out to be well there'd be a few years because there are erm some celebrities have in a same sex marriage and they've adopted and there's been pictures of them so luck then yeah yeah oh yeah he was like institutionalised it's interesting is it's okay to do that? yeah before
    \item 99. again i mean i spoke to him last week so that's fine for old time's sake handsome erm your man so weird maybe he'll lose god one day and realise that his marriage is a sham based around yeah while he's actually been in love with me for twenty years yeah exactly maybe how old are you? all i'm gonna say is before he got married to like
    \item 101. \#\#esar's background from what we right now know oh is that right? but er i didn't know yeah he's he was caught erm he well he had troubles with his marriage and i think that folded and then he was caught urinating on a g on a guy outside a nightclub you know and monty? that's amazing yes yeah but that's not the image that
    \item 103. wanted the boys just to be kind of normal i didn't want to sit round go oh mm because we had a chap ninety - three years you know? fifty - odd years of happy marriage yeah lovely career yeah yeah family home great children grandchildren mm nothing really to be sad about no nothing no no at all nothing to be sad no lived completely independently till the end mm you know so absolutely yeah
    \item 107. gave a beard? oh my god that's hilarious but now i'm taking it off her maybe doesn't fancy when she's got a beard and that's why no their marriage has gone wrong and er although i'll probably have to look after all the i took i took the beard off i'll probably have to look after the three children i need to give them pocket
    \item 108. dutch though erm he was a difficult man i wouldn't like to have lived with him god how long were they together for? oh for years ages? ages? yeah that was her whole marriage wasn't it? yeah it wasn't like she got married to someone oh no no that was her whole marriage and junior was the only one who just she wanted to leave him years ago but
    \item 109. together for? oh for years ages? ages? yeah that was her whole marriage wasn't it? yeah it wasn't like she got married to someone oh no no that was her whole marriage and junior was the only one who just she wanted to leave him years ago but she stayed together oh she should have done then she should have done she should have done of course he came here mm it
    \item 128. guy was saying he was sort of so he was so i th so he said a lot of people say like oh my family was absolutely fine until my dad walked out or my um you know my marriage was really was great until i found out my wife was being unfaithful and there there's lots of these kind of um so p people have like a big disaster in the family it just
    \item 129. ? he is nice yeah he does doesn't he? i think he has a wife and kids erm you know that he that he's like not l that long ago you know their marriage broke or whatever erm he's a nice guy he was so interested to talk to us like and to talk to about didn't he study classics or something? no i don't er
    \item 152. twins were by surrogate but she had one i think originally oh i see with matthew broderick mm they're still together aren't they? yeah after like twenty - odd years of marriage i think they had a bit of a rough patch but i'm talking about them like i know them personally they had a bit of a rough patch but they got through it in the end they got
    \item 163. just shows doesn't it? how things can change long long time ago and we were laughing at school one day about the fact that finishes six months before and this was long before any of their marriage problems i'm going back years and we were laughing about it and he said oh yes and i'll be standing on the corner of waving him off to work oh? yeah waving him off to
    \item 166. the boys are getting older now i've got myself in a life sentence which i don't have to have it's time a life sentence? yeah that's how he described his marriage that's a shame isn't it? i know and he said it's time now that i had get release something for me a bit of love and affection and he said and i thought
    \item 167. of this happened and he said it's made me realise that probably i'm approaching it completely wrong that actually what i shouldn't be doing is holding the family together and living in a marriage that's miserable i should be addressing the problem and sorting them out and making plans to split up really this is all i shouldn't be saying all of this oh well no well i'm
    \item 168. to be honest yeah well i think i mean if that was your husband or something you'd tell someone well leave your husband wouldn't you? you wouldn't say continue in a violent marriage no no i mean maybe in times gone by people would put up with it yeah but nowadays you know people would say you need to call the police yeah this person is dangerous what was he? what did
\end{itemize}

Cluster 2
\begin{itemize}  \setlength\itemsep{0em}
    \item 5. you he didn't do enough's right he'd been trying for like two ye he tried ttwo more than two years he tried for guns laws he's been trying for gay marriage for longer he's been trying oobama if you ever hear this but it's like they he'd been trying for obama care i know but it's not like it's
    \item 26. it's fair to come out and put down like gay people campaign agai people s you know people spend big big money to try and stop homosexuality? that's oh to try and get marriage illegal yeah religious and non religious people spend big money to do things like that i dunno if you wanna was after the american legalised same sex marriage to our fellow americans just tell me if i '
    \item 27. ? that's oh to try and get marriage illegal yeah religious and non religious people spend big money to do things like that i dunno if you wanna was after the american legalised same sex marriage to our fellow americans just tell me if i'm getting boring hey there it's two of your it's two of your brothers who are writing to you about the supreme court decision to legal
    \item 28. americans just tell me if i'm getting boring hey there it's two of your it's two of your brothers who are writing to you about the supreme court decision to legalise gay marriage the good news is that a whopping forty - two percent of you muslims support marriage equality as do both of our muslim elected officials in the united states congress one even serves as vice chair of the lgbt
    \item 29. your it's two of your brothers who are writing to you about the supreme court decision to legalise gay marriage the good news is that a whopping forty - two percent of you muslims support marriage equality as do both of our muslim elected officials in the united states congress one even serves as vice chair of the lgbt equality caucus there are many faithful gay and lesbian muslims in the us and we love and
    \item 30. fast during ramanam and swap turkey bacon on your blt all in an attempt to establish a firm muslim identity in a non muslim country but now that s now but now but now that same sex marriage is legal in america it's shaking up your faith you're afraid of the future and what this could mean for your kids you recognise this growing acceptance of gay rights but personally you just can '
    \item 31. \#\#s so that we can easily identify us as a muslim we shouldn't be per we shouldn't be perpetuating our marginalisation by marginalising others rejecting the s right to same sex marriage but then expecting empathy for our community struggle is hypocritical think about the way er people look at your hijabi sisters what i said earlier or your bearded brother when they walk through the
    \item 44. it's like when something happens in the law and then three months later there'll be a natural disaster and then someone religious will say it's this hurricane is because they legalised gay marriage three months ago it's horrible but it's it's horrible they they s and they've got children as young as six holding holding a erm sign that says god hates fags
    \item 46. kids and counting? yeah the show was cancelled after josh oh fucking hell oh what? fucking hell oh that is vile oh oh god oh right touring the country giving talks on the sanctity of marriage and campaigning against same sex marriage what the fuck? fucking hell wow he sounds like an absolute dd what's mm that ashley madison thing? it's like erm er infidelity so you
    \item 47. show was cancelled after josh oh fucking hell oh what? fucking hell oh that is vile oh oh god oh right touring the country giving talks on the sanctity of marriage and campaigning against same sex marriage what the fuck? fucking hell wow he sounds like an absolute dd what's mm that ashley madison thing? it's like erm er infidelity so you go on it to like er
    \item 58. ' t i can't he's just walking his head's like this high just bobbing along oh my god but yeah he was pretty big today we discussed er a lot about gay marriage fundamentalism do you know who would've love this? yeah no shall we shall we use his code name for the podcast? well micro penis? no i like squadcast it's nice squad
    \item 94. they believe in having numerous er children cos that gives better better gifts to god or something i don't know and they were all going on a march in madrid that weekend mhm against gay marriage because gay marriage was just about to be legalized in spain mhm and there was a a girl in class who was a lesbian and quite openly so and she came to class with three of her friends
    \item 95. having numerous er children cos that gives better better gifts to god or something i don't know and they were all going on a march in madrid that weekend mhm against gay marriage because gay marriage was just about to be legalized in spain mhm and there was a a girl in class who was a lesbian and quite openly so and she came to class with three of her friends from school so
    \item 117. so mobilised about an individual issue it's a pity what with paedophile priests aids third world poverty yeah et cetera that they w of all the things to get excited about is gay marriage yeah a bit more interesting a bit more worthwhile so do you honestly want to punish people you don't like or help people? you're catholics mm sorry let me be more accurate you '
    \item 155. mm so we'll see mm yeah but it was you know it was just such a shock and then he started a sentence that was something like oh i just don't get that about gay marriage and me and were like no no no no no no oh he's like i just don't believe that people you know are so against it and i was like thank fuck for that i was
    \item 156. this man i can't live with this man i'm sorry yeah we were like er this was sort of after the ukip revel er revelation yeah mind you david cameron voted for equal marriage there's still big problems it's a big issue in northern ireland still erm mm mm with er the dup vetoing it all the stormont executive has said yes let's have
    \item 157. s still big problems it's a big issue in northern ireland still erm mm mm with er the dup vetoing it all the stormont executive has said yes let's have equal marriage and the dup er have this petition of concern mm which is a a legal thing that they can use to block legislation from being passed if they believe that it's gonna unfairly disadvantage either the
    \item 162. s just too much it's too much you have to say something for the cambridge but yeah miss interesting views yeah i'd say that interesting views are like i i assume miss miss supports gay marriage she supports well she supports gay rights erm she supports i don't think she's miss i think she's cool gay rights at least i i think maybe just she support gay rights i
    \item 169. country mm who she goes out to on the tram erm and teaches english to and this woman apparently is ultra ultra conservative uhu and also's facebook friend and had put some pro - gay marriage campaign thing or changed yeah the rainbow picture oh that's what it was yeah she put the rainbow picture on in support of the american supreme court mm erm saying yes gay marriage across the country or
    \item 170. pro - gay marriage campaign thing or changed yeah the rainbow picture oh that's what it was yeah she put the rainbow picture on in support of the american supreme court mm erm saying yes gay marriage across the country or marriage for whoever wants it across the country and her student picked her up on that and started quizzing her on her own beliefs erm and said look i'm your teacher i
    \item 171. thing or changed yeah the rainbow picture oh that's what it was yeah she put the rainbow picture on in support of the american supreme court mm erm saying yes gay marriage across the country or marriage for whoever wants it across the country and her student picked her up on that and started quizzing her on her own beliefs erm and said look i'm your teacher i'm here to talk
    \item 172. so that all the priests who are pra um who are celebrating mass today were gonna read out a petition to ask their congregation to turn against the government's plan to introdu introduce same sex marriage the government want to introduce it make it law by twenty fifteen yeah civil partnerships but the church is absolutely against it couldn't believe it even now oh my god i'm gonna speak to my dad
\end{itemize}

Cluster 3
\begin{itemize}  \setlength\itemsep{0em}
    \item 158. big petitions trying to get their petition overturned to bring it in to to law that still hasn't happened mm mm mm i do wonder you know if it does come in if my legal rights of marriage w s might get backdated because and i had a civil partnership in northern ireland mm mm so i don't have equal rights you know financially and that's going to affect me in later 
\end{itemize}

Cluster 4:

\begin{itemize}  \setlength\itemsep{0em}
    \item 0. de montaigne he's a philosopher isn't he? a french philosopher or something oh is he? mm about the the thing about a friendship the difference between a oh friendship and a marriage right i like it i mean i looked it up and tried to read it and er you'll probably be better at it than i am but it was re really hard going with english mm and er
    \item 11. day? yes it was actually yeah a nice day how long were they together before they actually got married? erm er that i don't know they were cos people seem to dive into marriage pretty quickly don't they? yes yeah exactly might be half of the problem i mean there was one there's there was a song in the charts that that you kind of does sum it up
    \item 13. and the south coast and that kind of thing but not it's not just about women but he he does think and also he's sort of saying you know m the whole original rhetoric of marriage being with one person was when people were dying in their fifties and sixties and that made sense you know yeah you sort of got married when you were twenty yeah and then you're gonna be dead by
    \item 21. about a send - off before you go bloody out there? you can't god hold on he's only just got back he's not going anywhere no but he's talking about marriage we're talking about the marriage thing and he doesn't want we may not even get married if we if we to go there you have to yeah but if we get a job that's
    \item 22. go bloody out there? you can't god hold on he's only just got back he's not going anywhere no but he's talking about marriage we're talking about the marriage thing and he doesn't want we may not even get married if we if we to go there you have to yeah but if we get a job that's not there we're not explicitly
    \item 43. jane the virgin you watched jane the virgin? isn't it amazing? what the hell's that? awkward basically there's like she promises to is to not to not have sex till marriage yeah she does and then and when she gets to have a when she goes to have a smear test she accidentally gets artificially inseminated oh ch no mm hm awkward and she's engaged
    \item 49. okay i can show you the picture of me proposing so erm i think i've swapped over so if we get a job in the middle east we'll sign the paper next may no marriage nothing just forty pounds sign a piece of paper so are you doing anything with mum? dad? or anything of that nature? some sort of cultural blessing? or? well we'll have to have
    \item 51. re married but they're not no you know they've had a son they live together do everything but she goes we just don't don't need it don't believe in marriage don't wanna get married no right and i think after those many years it's like well why do it now? and they almost feel like maybe it's like jinxed or something it
    \item 54. marry though are you? well yeah because if you are i'll be i wouldn't want to kiss i love that we're working this out like on eggheads yeah yeah would be marriage material even if the attraction's not there he would be he would be a nice husband wouldn't he? let's be serious you would rather marry than i'd rather marry none of
    \item 63. if i could have yeah? cos i think gays should have the same right to marry as anyone else yeah cos marrying can be a really good thing for someone yeah why do you think marriage would be a good thing for the people? erm it's kind of like it's kind of like you you share a bond with each other and you really like love each other and you
    \item 64. you necessarily have to have a good job or a good house you can just have i it doesn't really matter i think as long as you are committed to your partner then that's what marriage is about and not about a good house or a good good like not about having a rich house or anything just about commitment and just loving each other do you think they might need sa stability though? i
    \item 65. ? i think stability would be a very a very big factor in it i think it'd be hard to marry if you didn't but i don't think iid let anything stopping marriage i've ba a lot of people getting married and then they're they want kids so they want to bring up kids in a stable way don't they? yeah they're not kids
    \item 69. ? that's average of what people spend oh yeah that's the average of what people spend it can go down to about half of that i was gonna say it it's like the marriage thing you can get married for two hundred pounds but they quote fifteen thousand pounds to get married that is the same as these quotes on funerals but seriously they can blow lots yeah you don't need
    \item 75. you have one t? damn it er i could still do that er it's tricky if you look at that what word do you see immediately? rain er ira oh air okay yeah i see marriage but i'm missing marriage? i'm missing a few letters but don't you just see marriage? is that a freudian slip there? no yeah i can see that it sounds like
    \item 76. it er i could still do that er it's tricky if you look at that what word do you see immediately? rain er ira oh air okay yeah i see marriage but i'm missing marriage? i'm missing a few letters but don't you just see marriage? is that a freudian slip there? no yeah i can see that it sounds like a freudian slip erm
    \item 77. that what word do you see immediately? rain er ira oh air okay yeah i see marriage but i'm missing marriage? i'm missing a few letters but don't you just see marriage? is that a freudian slip there? no yeah i can see that it sounds like a freudian slip erm or carriage perhaps boring boring hey there was a carriage in the chelsea flower show was
    \item 78. ? oh it was cool i watched one episode i think there were only three episodes ever i couldn't believe they were getting married oh they did but i didn't but you can nullify marriage within like yeah two months can't you? yeah exactly so that's what they did yeah so they had five weeks oh okay to and then if they didn't like each other yeah it
    \item 80. how they ended it oh i think that was the last one in the series wow but i guess they'll do another series it is quite cool because yeah you know in the middle east like arranged marriage is still the way it's done but it's not erm blind like that normally okay like they do normally meet like normally it's the parents and the brothers and the sisters are out
    \item 92. they didn't wanna make a change when they got married mm but after being married for a while you know like me er they decided they would um or they're in a relationship leading to marriage we've got a couple er like that there's a hungarian guy and a um er an asian er woman who are getting married fairly soon and and er mm the they want so the they they
    \item 115. hell was a very appropriate response mm my favourite thing is that he going on about so they had to they had to get something to focus their argument around and what it is is that they focus around marriage should be marriage is for creating a stable environment in which to bring up children right so infertile couples shouldn't be allowed to get married then fuck off and die you bigot sorry mm it
    \item 116. very appropriate response mm my favourite thing is that he going on about so they had to they had to get something to focus their argument around and what it is is that they focus around marriage should be marriage is for creating a stable environment in which to bring up children right so infertile couples shouldn't be allowed to get married then fuck off and die you bigot sorry mm it's alright
    \item 118. okay my life was a lie okay it is with an s they lie they do ah do you feel like getting married to? yeah absolutely i mean to be honest i don't really agree with marriage anyway so yeah for me like yeah i'm starting to think that you know? like that marriage might not be and it's like a considered a classic you know the way pathway yeah but that
    \item 119. like getting married to? yeah absolutely i mean to be honest i don't really agree with marriage anyway so yeah for me like yeah i'm starting to think that you know? like that marriage might not be and it's like a considered a classic you know the way pathway yeah but that's that's the problem with it but it's the classic pathway like it's
    \item 120. ' t afford it a wedding a house and a baby shit i'm not gonna do that am i? no think the baby'll come first remember you said you wanna have a baby in marriage didn't you? i understood it was okay to have a baby out of marriage okay so but then you'll also have to wait till you've got enough time to lose the baby fat
    \item 121. do that am i? no think the baby'll come first remember you said you wanna have a baby in marriage didn't you? i understood it was okay to have a baby out of marriage okay so but then you'll also have to wait till you've got enough time to lose the baby fat true to get married to get married no one wants to be fat on their wedding photos
    \item 122. so but then you'll also have to wait till you've got enough time to lose the baby fat true to get married to get married no one wants to be fat on their wedding photos marriage first and obviously you don't want the baby being too young because then they'll like oh i don't want i don't want babies in wedding pictures that's just yeah gay
    \item 123. reason that we're getting engaged really it was we were talking before and like we we've spoken so sorry i've got it no no it's fine we've spoken about marriage quite a bit and mm this time it actually got to the point where we were talking about practicalities of marriage mm mm who would come where we'd do it how much it would cost when we
    \item 124. sorry i've got it no no it's fine we've spoken about marriage quite a bit and mm this time it actually got to the point where we were talking about practicalities of marriage mm mm who would come where we'd do it how much it would cost when we'd do it and at this point i kind of thought that's really not very romantic is it?
    \item 127. is it's to like keep the church going it's more like we've been this pra yeah i agree yeah and it's we've been so pragmatic about marriage that you might not even change your name because it's only forty pounds to get married it's a hundred pounds to change your passport so like it triples the price of the wedding just to
    \item 136. home mm but she's not she's just like sitting in the living room watching my tv not doing any work but having her laptop on her lap taking up the entire downstairs this is what marriage is like you know i'm not married to her it feels like it it fucking annoys me i'm just like go away and she's like just wrap it up or eat it don
    \item 148. if i do want to get married soon i'm probably gonna have to back down on that cos i'm not going to convince him otherwise what is it you want about a catholic church marriage? just because i dunno it's more of childhood thing like i just wanted to get married in the church i've always gone to and i still go to when i go back to hereford
    \item 164. unloved one yeah tut ah dear don't love her so she's showered with money tut definitely tut right so when are you planning to get married? like i said marriage isn't er something on my topic isn't it? it's not my cup of tea no so you're gonna stay single for the rest of your life? yeah you seem keen
    \item 165. that's the next thing oh no no i oh okay why? is that what you believe? no no no as soon as you're married have kids like i said pop pop pop pop marriage isn't my cup of tea no what the hell? what's with the sound effects? so what is it? you're gonna be like a baby machine? pop pop pop no that
    \item 175. of good morals and they're yeah they have they are nice people but some things that i don't agree with they say well we're really good we don't have sex before marriage but then they get married at like fifteen so yeah yeah well that's probably s that can be er younger than the legal age anyway so it doesn't really count yeah yeah and the way they
\end{itemize}

Cluster 5
\begin{itemize}  \setlength\itemsep{0em}
    \item 8. that i'm glad that he was not suggesting that you beat your actual wife no his actual wife beats him just your pretend wife yeah it's his fa it's his fake loveless marriage aw it's his faithless violent marriage it might be a fake lo m fake marriage but it's not loveless aw that's the saddest sentence i've ever heard yeah
    \item 9. suggesting that you beat your actual wife no his actual wife beats him just your pretend wife yeah it's his fa it's his fake loveless marriage aw it's his faithless violent marriage it might be a fake lo m fake marriage but it's not loveless aw that's the saddest sentence i've ever heard yeah aw poor with his deep - filled ham sandwiches
    \item 10. actual wife beats him just your pretend wife yeah it's his fa it's his fake loveless marriage aw it's his faithless violent marriage it might be a fake lo m fake marriage but it's not loveless aw that's the saddest sentence i've ever heard yeah aw poor with his deep - filled ham sandwiches deep sorry but there's a massive inn
    \item 33. like you know he's best off without her but they're married couldn't deal with tension so they'd been married for a year and a half did he not realise before the marriage that she was a nutcase? you know what? i think she just really liked the idea of being married mm and they have this like fairytale hollywood like marriage like they got married it was you
    \item 34. did he not realise before the marriage that she was a nutcase? you know what? i think she just really liked the idea of being married mm and they have this like fairytale hollywood like marriage like they got married it was you know they pumped all their life savings i mean that wedding cost at least a hundred and twenty grand oh my days it it was the you know it was the works so
    \item 45. mar to be married legally to more than one person because as soon as you if you are already legally married to someone and you try and legally marry someone else in the eyes of the law that second marriage is invalid because you're already legally married so it's actually impossible to be legally to commit the offense of bigamy oh go no thanks you made me jump boom no no you looked at it
    \item 70. on the immediate family as well yep so so for example oh so the child is indirectly affecting itself? yeah by affecting those around it? exactly so all good gifts around us mm sent so a bad marriage from heaven above a bad marriage can influence the child but a fussy baby can also create frictions in a marriage wow so oh and then vice versa just keeps on going oh dear so yeah so that
    \item 71. yep so so for example oh so the child is indirectly affecting itself? yeah by affecting those around it? exactly so all good gifts around us mm sent so a bad marriage from heaven above a bad marriage can influence the child but a fussy baby can also create frictions in a marriage wow so oh and then vice versa just keeps on going oh dear so yeah so that's the main oh dear
    \item 72. those around it? exactly so all good gifts around us mm sent so a bad marriage from heaven above a bad marriage can influence the child but a fussy baby can also create frictions in a marriage wow so oh and then vice versa just keeps on going oh dear so yeah so that's the main oh dear oh dear principle of the microsystem mm then there's the mmacr
    \item 73. together which were daughters she probably wanted a son he probably wanted a son with her i meant it must have been debated on he would have liked to have a boy that never happened he then had another marriage that didn't work and then he had a marriage again and had a child erm so for him for for them they they feel like feel like our family is is it doesn
    \item 74. wanted a son with her i meant it must have been debated on he would have liked to have a boy that never happened he then had another marriage that didn't work and then he had a marriage again and had a child erm so for him for for them they they feel like feel like our family is is it doesn't it doesn't necessarily cement the families together no it doesn '
    \item 81. oh and yeah but she's still allowed to say no she can oh right even when they meet she could say it's i don't like him so it's not a forced marriage find another one yeah yeah and then they'll just bring another one and that can go on for ages yeah but what's a bit nasty in the gulf is it's linked with money i
    \item 113. about the pragmatics the practicality of getting married to the point where we're actually talking about who would come and where we'd do it and we wouldn't get a marriage we'd just sign a piece of it wouldn't be a wedding just a marriage we'd just sign a piece of paper yeah you'd just do the thing and i was sitting there
    \item 114. re actually talking about who would come and where we'd do it and we wouldn't get a marriage we'd just sign a piece of it wouldn't be a wedding just a marriage we'd just sign a piece of paper yeah you'd just do the thing and i was sitting there thinking that's not very romantic is it? if she we just go and sign a
    \item 143. divorce in those days i think it took a long time yeah and i can't find him dead so anyway i've actually written it in i think so i think this a bigamist marriage oh god so more recently when the newspapers have come out it's in the newspaper it is you're right and it's it's quite hilarious because it was a biga a big
    \item 144. more recently when the newspapers have come out it's in the newspaper it is you're right and it's it's quite hilarious because it was a biga a bigamist marriage erm a bigamous marriage and we if you read all the detail she was quite sharp she knew what she was doing say but she actually was found not guilty because he lived he lived about ten
    \item 145. out it's in the newspaper it is you're right and it's it's quite hilarious because it was a biga a bigamist marriage erm a bigamous marriage and we if you read all the detail she was quite sharp she knew what she was doing say but she actually was found not guilty because he lived he lived about ten miles away from her and they '
    \item 146. so it sort of to me it sort of revolved a a lot around you know about this relationship with this man you know the the daughter father relationship and the as my mother kept putting it failed marriage you know the way she'd phrase it yes yes failed marriage so erm so really seeing that all guns were pointed at you know erm mm yes and er and that's the way i
    \item 147. lot around you know about this relationship with this man you know the the daughter father relationship and the as my mother kept putting it failed marriage you know the way she'd phrase it yes yes failed marriage so erm so really seeing that all guns were pointed at you know erm mm yes and er and that's the way i've been brought up you know mm yes but as i say
    \item 150. mental institution yes so conan doyle is watson and he he's married but he begins an affair conan doyle does? yes er sort of inadvertently but there's it's not his with this marriage? not exactly no sort of thing no? but plutonic marriage and he starts an affair his wife then develops a terminal illness mm and he supports her very faithfully through that while also seeing his mistress
    \item 151. married but he begins an affair conan doyle does? yes er sort of inadvertently but there's it's not his with this marriage? not exactly no sort of thing no? but plutonic marriage and he starts an affair his wife then develops a terminal illness mm and he supports her very faithfully through that while also seeing his mistress who he then marries after his wife's death mm but he
    \item 154. there could be between them but she had a lot of freedom as a married woman and did a lot of painting and became a painter in her own right so so possibly it it might have been a marriage of convenience i suppose but certain she had a lot of freedom to be an artist and and how did you come by the connection with your house? well somebody er somebody who's related to both of
\end{itemize}

Cluster 6
\begin{itemize}  \setlength\itemsep{0em}
    \item 104. get it deary me well the the first step was made but unfortunately the second second step has proved a hurdle which is the actual not drinking of wine sorry so i've spotted a performance of marriage of figaro and of all all the places no of all places it's in yeah sounds good so i'm thinking i should invite my mother ah that'll be nice okay my love you
    \item 105. ' m going get all of these pots in the dishwasher no don't worry just er just put hot water and soap in them and leave them in the sink heating these up so er marriage of figaro dad sing me a key theme right you've got all your dark wash is washed and dried thank you mum there's a beautiful aria on the cd that i bought from figaro
    \item 106. sure what to do about it anyway i've forgotten to write to him and now he's gone behind my back erm right an array of socks some of which could be yours is that marriage of figaro get clothes on mum's like come on get on with it there's a certain enthusiasm towards the pub visit i've noticed is this after half a day without a glass of
    \item 133. i was just about to turn it off and suddenly this lovely aria appeared oh right two the two sopranos er mm who marries figaro and er the countess really lovely oh right were you listening to marriage of figaro rather than magic flute? marriage of figaro yes mm yes did i say magic flute? mm marriage of figaro only cos i was thinking cos i saw the magic flute in
    \item 134. suddenly this lovely aria appeared oh right two the two sopranos er mm who marries figaro and er the countess really lovely oh right were you listening to marriage of figaro rather than magic flute? marriage of figaro yes mm yes did i say magic flute? mm marriage of figaro only cos i was thinking cos i saw the magic flute in plymouth mm when glyndebourne were touring
    \item 135. who marries figaro and er the countess really lovely oh right were you listening to marriage of figaro rather than magic flute? marriage of figaro yes mm yes did i say magic flute? mm marriage of figaro only cos i was thinking cos i saw the magic flute in plymouth mm when glyndebourne were touring mm i really liked it mm yeah the um you had it on vhs
\end{itemize}

Cluster 7
\begin{itemize}  \setlength\itemsep{0em}
    \item 17. make excuses to my girlfriend as to why what was and the rest is history was there a point then there where you kind of both decided that you're gonna break up you know finish her first marriage or was this something she decided or oh well i she probably erm yeah once i was on the scene erm i think it was found out about it did you get any problems from? no i
    \item 18. quite a difference between the two of them yeah yeah yes but he's got a whole other family's got erm a half - sister stepsister half - sister half step is is through marriage but no blood no she's in her forties with like children oh good lord she doesn't see them very often i think her dad's calmed down quite a bit as he's
    \item 55. he come into it? had the same dad as your mum but i don't understand i think he is related to the family not by oh he's blood line but i think maybe through marriage or something oh so he is he's related oh that makes it a bit less weird yeah i thought i cos i didn't understand what was going on and i just thought that i don
    \item 61. on a saturday morning i know this is really short notice but is there any chance you could babysit tonight? it's all of them cos she's got three children from her first marriage and little from her and and erm when has looked after it it's only been she's looked after right cos if it's all of them she tends to get her mum so
    \item 84. thing so this is your mum's dad mm who had your mum is your mum born first or after they were my mum was born first mm yeah and so mum er my gran then the second marriage is the other ye well actually that's the thing at the time that was what happened was probably why they never met was because though my fa my grandfather married my mother er grandmother well that was her
    \item 85. actually that's the thing at the time that was what happened was probably why they never met was because though my fa my grandfather married my mother er grandmother well that was her name yeah became through marriage and then they got divorced when my mum was quite about er erm young maybe five - ish around that mark oh right so does oh you don't know did your mum remember him? you
    \item 87. it's not perfect but it's it certainly seems to be fairly yeah solid i've yeah well i i'm happy to come over and oh dear's cousin's by marriage you i don't know if you remember and that were in erm they were at our wedding do last year no? well anyway they moved to bulgaria they've bought a house in bulgaria really
    \item 100. got four brothers all these brothers okay wow yeah erm i def i was gonna say i definitely thought he had an older brother he's got and who were his da his dad's first marriage and then he's got who was his mum and dad's okay so yeah he's got four brothers and then obviously he's got shit loads of family and like cousins and stuff like
    \item 110. she was sleeping with her boss exactly oh disgusting really not at all have lived er are they still together your parents? no no have they remarried? yeah both of them have yeah which is nice second marriage? or have they been married again? no only both second marriage my mother's on her third marriage now is she? yeah ridiculous is he nice? yeah he is ni i've known him
    \item 111. have lived er are they still together your parents? no no have they remarried? yeah both of them have yeah which is nice second marriage? or have they been married again? no only both second marriage my mother's on her third marriage now is she? yeah ridiculous is he nice? yeah he is ni i've known him now for a little while they divorced when i was seventeen and she
    \item 112. parents? no no have they remarried? yeah both of them have yeah which is nice second marriage? or have they been married again? no only both second marriage my mother's on her third marriage now is she? yeah ridiculous is he nice? yeah he is ni i've known him now for a little while they divorced when i was seventeen and she'd already been married before she met
\end{itemize}

Cluster 8
\begin{itemize}  \setlength\itemsep{0em}
    \item 2. now oh did you see the one jennifer lawrence wore? yeah emma stone emma st stone's one was beautiful it was like oh did you see the one jennifer cat's eyes yeah she does marriage and i'd have bridesmaids who are wearing that you wanted it would make you look like your tuxes but they have like a topshop unique line and just what i'll do if
    \item 4. dates go or is it just me? i don't know mm maybe if they cooked it and then you're like oh you're an amazing cook no and and frankly i i see marriage as simply eh you know for material benefits so eh you can cook you'll do is my motto em oh you're very cynical i wonder how well sarcasm will show up like every line has a
    \item 14. how much money you've got and and i think that's really worrying and there's still i so sort of speak to girls here at david who all they're after is the marriage they don't mm mm and it doesn't matter who it's to as long as they're rich they don't care about the person how shallow what a shame it's
    \item 15. re like both like girls who are quite into like er j wwwwere friends like w like quite a few years ago we used to have some things in common but they're very much like marriage children you know like they're on that path and they're and they and erm i took the timing quite lolzy so like the waitress was coming over tto take our order and
    \item 24. no need to worry reassuring yeah reassuring to know thanks for that but um er and he was like but then he kept making comments like not just muslims that are with their cousin or sleep with cousins but marriage to their cousins right um and i'm probably reading too much into it but then he was asking about how many people i'd slept with and stuff mm his number's off the scale uh
    \item 25. okay yeah i'm not in favour of the idea in general i think it's er it's a bit of a strange thing to do yeah i'm not it's a marriage of slavery actually branding of course is a a marking of the skin for you know bad reasons yes but then you have erm but then you have them tribal markings don't you like you like i
    \item 36. yeah what a crazy woman yeah but it's just you know it screws up the whole family's life doesn't it? because yeah i bet he's regretting the whole marriage thing now yeah he's a bit jaded like when we went to that party last weekend he was just like i'm going for an hour saying hello to sonia i was like ah that '
    \item 50. ' t traditional no wwe're erm where the status quo is the other way we may not even change her name like if we get married because changing her name would triple the cost of the marriage see it's it's only oh forty pound to sign it down the registry office but it's a hundred and twenty quid to get a new passport yeah a lot of people won '
    \item 56. each other yeah a great deal you know erm i just think in life you have to have yeah realistic yeah hopes expectations yeah yeah i go i agree with that i think you do and i know marriage is a plain sailing one and i'm always being told that by my mum so that was good yeah yeah yeah exactly so she kind of said forewarned me and said well you know what?
    \item 59. one one asian named kid one white named kid yeah you you yeah you flip a coin which one's the asian kid i think that's a major discussion point though within a couple in a marriage what? child names baby names especially between i don't know sometime sometimes it's different sometime they let one of them take first one like okay so really wanted to call him because his grandfather him
    \item 62. invented in shakespeare's day i can't hear a word no no horsepower was though did you know that they found erm marijuana in erm? marijuana? marijuana oh i thought you said marriage he was taking the mickey i know he was but i didn't understand it took me a minute to work it out so yeah so they found marijumari oh marijuana in shakespeare's like
    \item 66. can pay for it or at the definitely not gonna happen the coach and twelve horses there's there's a friend of mine who reckons he can prove that the more you spend on the marriage the less likely you are to survive your future life as a as a married couple he reckons well why? looking ar well just looking around his friends he seems that seems to be a correlation and to
    \item 67. many times do you hear do you hear of you know oh you know they how can they afford to buy a house and all the rest of it but they've just had a twenty thousand pound marriage now which is actually better? twenty thousand pounds on one day or ten thousand pounds and a deposit for a house? well don't confuse mm correlation and this this modern with causation what sorry
    \item 68. yeah it could be that people who are bit more showy mm are less arguably authentic yeah oh no could easily be anything and want to spend more yeah but there's nothing more authentic than a marriage cos you get to yeah know someone warts and all yeah don't you? oh yeah oh yeah say say nothing oh yeah warts and all eh? warts yeah warts and war
    \item 82. they can kind of be with probably the person they've loved all their life you know they've got someone else in mind in the back of their mind until they've done for the marriage then then they're not gonna be allowed to marry them yeah i reckon that's what and get the money really happens and the yeah that's a shame yeah yeah and there's a
    \item 88. the surname then you could get away with it but it w still wouldn't be legal would it? and if she sh we can change our names when we get married but the cost of the marriage at the the place on a weekday is like forty pounds cost of a new passport is like a hundred and twenty pounds it's literally like three times the yeah but don't forget you're
    \item 102. little older than that no no was sixteen? yeah or eighteen yeah actually i think she was seventeen well maybe yeah i th yeah tha you're right but i think mm was nineteen i mean a marriage you almost well that's a bit like rather did you meet them? did you come to that day? no no mm i mean they married you know when they were tots they spent their first
    \item 125. we find a joke that in an islamic country the fact that we've been together for nine years counts for nothing mm mm but if we got married if we are married for a month a christian marriage with a christian marriage that apparently holds some weight but yeah it is it's just how it is it's to like keep the church going it's more like we've been this pr
    \item 126. that in an islamic country the fact that we've been together for nine years counts for nothing mm mm but if we got married if we are married for a month a christian marriage with a christian marriage that apparently holds some weight but yeah it is it's just how it is it's to like keep the church going it's more like we've been this pra yeah i agree
    \item 130. things i totally understand it and i haven't and i like no no mm misconception it's just having children massively complicates the situation so mm mm than the actual marriage bit yeah but i mean mm if you remove the children divorce it doesn't really mean anything does it? no no hence my first it was nothing and it was yeah it only hurts you and the
    \item 132. is yeah they all tend to be around here don't they? it's just you know it's just typical four five metres wide victorian type houses all around mm so you mentioned the marriage word have you been uh inspired by and getting engaged? no no no no long way to go yet long way to go yet but having said that i am thirty - five this year are you? that
    \item 141. she was she was known as yes yeah never though it is so when was your mother born? in nineteen forty - two oh forty - two i see yeah yeah right so that and then that is marriage to so fifty - four fifty - four five six seven eight nine fifty fifty - one two three four yeah she was quite a bit younger oh there you are so i was nine when he was yeah you
    \item 176. her star sign's was turquoise i knew most of them but i know mine's emerald june is pearls i think july is rubies you do know all the erm presents for years of marriage? yeah and s i know the main ones you've had most of them but i don't know all of them cos there's so many i mean you start off and you '
\end{itemize}

Cluster 9
\begin{itemize}  \setlength\itemsep{0em}
    \item 1. mm um recruiting more radical muslims mm to follow this guy's name i don't know cos i wasn't really paying a lot of attention mm um and you know force girls into marriage and really really extreme views mm but the the kind of idea of a conversion rate between the three percent muslim mm and the percentage of those that are extreme mm was really very high whereas the he was arguing
    \item 12. this then you don't get deported mm well his his grandfather the celebrity's great grandfather was english and his name was yates yeah but the the trib the funeral erm decided that the marriage that erm was had taken place between erm the his his father's parents sorry grandparents wasn't official so therefore they didn't recognise it fortunately cos his mother had got british she
    \item 35. you know it was the works so they had like a beautiful ceremony and they had like this amazing reception in um park lane right so they were she was pumped by the idea of the wedding not the marriage? absolutely and they you know they had all these like you know folk dancers in between he came in on a horse you know it was like the full - on shebang when did they meet? how
    \item 37. hour saying hello to sonia i was like ah that's so sweet so much fun then i remembered hold on a minute he's you know quite tender actually and we're going to a marriage celebration so don't rub it in oh is that the part?'s you know that leaving party oh i thought ah it was leaving party? oh cos she was leaving to go and get
    \item 39. because it's a different office oh no and they can't oh they're not can't they they can't which means access your i will need your london copies? my marriage certificate your passport your marriage certificate all of that all of those the originals? no certified copies probably originals and certif erm and i had to do that three colour certified copies yes because i had
    \item 40. different office oh no and they can't oh they're not can't they they can't which means access your i will need your london copies? my marriage certificate your passport your marriage certificate all of that all of those the originals? no certified copies probably originals and certif erm and i had to do that three colour certified copies yes because i had when we did it in
    \item 60. ' t you? i suppose so kind of that they have had bad luck yeah yeah and you know bad luck at yeah birth and and yeah you do in in the early years and bad luck with marriage yeah i don't know what'll happen but anyway so it's all it's it's usual messy self up there but she's going away cos i invited her has
    \item 79. matched three oh so that was but they they had to be a really high yeah success rate they went really really high to to make it work and then of those three couples one pulled out before the marriage oh cos the girl kind of got cold feet and her family didn't approve so she was ththat was out when she when she met him? no th no they weren't allowed
    \item 86. ' t there no it was just the girls they're they are a weird couple though because i think they are like they're moving at very different spa paces like's ready for like marriage kids house the lot what? and's just like no way like doesn't wanna buy a house with her yet doesn't wanna move in with her erm like she was goi she
    \item 89. here was cos her cousin is only down in fareham alright okay so she's only half an hour's drive away mm and she er er had sent me a l copy of the marriage certificates of my grandfather and we knew about my mother's father's side but not my mother's mother's side so i just put in put in her name and and the records came
    \item 90. ve been he could've made a lot of money he he could've done i guess so he possibly did and like the american or something? mm he could've done and on this marriage certificate oh his daughter's marriage to erm my mother's father erm he's described as of independent means oh but the he made a fortune out of it erm well plaus
    \item 91. a lot of money he he could've done i guess so he possibly did and like the american or something? mm he could've done and on this marriage certificate oh his daughter's marriage to erm my mother's father erm he's described as of independent means oh but the he made a fortune out of it erm well plausibly he has but in the same
    \item 93. memberships that are a requirement mm of the job role mm that you're seeking mm so yeah mm so it's it's kind of er it's finding the right kind of marriage mm between the two mm really isn't it yeah okay do you think we've said enough about cvs? i think we probably have probably bored them with cvs yes let's stop
    \item 97. at oh and then yeah do you know she still has a reunion with her digs friends there are nine of them right? they're all girls and they're all still on their first marriage wow mm heaven for you it's interesting that isn't it? it is i'm not sure what the statistic or it's some just women? no no it's both
    \item 98. oh by the time you're twenty - five like you'll be in a really serious relationship you would of bought a house you'll be talking about kids you'll be talking about marriage and it won't and it won't scare the shit out of you i'd be like fuck off yeah like i had no plans to be in that position by the time i was twenty
    \item 131. gonna be at different institute so mm it's we're not gonna see each other in the day no not at all i was saying to my friend some of my friends have asked about our marriage and stuff and like are you actually married and it kind of annoyed me a bit that like apparently in my dad's house so's yeah the doctor missus they're married now and somebody
    \item 137. they just said bring all your papers what are they? not a lot well i mean you you should but i've got my my pension things here and er yeah your i've got my marriage book yeah and i'll take er's er form i've got to say that er what? she's registered with the british government with the yeah yeah so you've got all
    \item 138. \#\#sque stuff and i was like oh why would you want the entire train to know this stuff? like one of the girls was talking about erm well they were both actually talking about erm arranged marriage no sending like nude photos to to boys and stuff no and one of them ccos basically one of them had clearly like clearly kind of broken up with a boy fairly recently and they'd had to
    \item 139. ? okay one of those things um it it is sort of square prism comes out at the top of a house um chimney dick van dyke it says uh oh no pass uh the male person at a marriage ceremony stop groom? oh never mind uh four? yeah yeah very good yeah coupon rice touch chimney okay uh sarah you're asking me no oh we go around this way now? you we always
    \item 140. nineteen forty - two that's my half - sister yeah oh yes i got that down yeah then this is his erm oh that's his birth certificate and and then that's their marriage certificate to gosh yeah yeah and here now that's interesting isn't it because then because my grandmother's so right okay so she must have changed her name around at that point i don
    \item 142. s how i go can go back you find out where she was born so then you've got her mother's name and her father's name and that's so this is their marriage certificate they were married at the congregation in and then their death certificates mm so that was a sub - acaroid haemorrhage acaronoid haemorrhage yeah i don
    \item 149. wasn't three hours we looked we all looked really good we really made an effort i think we did sexy witches oh back in our youth i know it's crazy everyone's talking about marriage and babies and all that mm who do you class as everyone? everyone every single person on this planet bet's not she wants to's i think it's just an awkward situation isn '
    \item 153. s nobody looking after on wednesday so wednesday i've got to spend six hours there cos no one else will look after it there and have his visit and which we have everything out burial register marriage register christening register service records yeah? and i've got to check everything and mmake sure everything's there on this the list of stuff the whole lot oh oh right just see if it
    \item 159. years until we die thirty years ah makes well you've done you've done the first of the three ms but obviously you remortgage and you move you've got mortgage marriage is coming and then there's just maternity after oh i was gonna i was trying to work out how it's murder it's murder that's the fourth and final m do you guys
    \item 160. ' s like heaheavily homophobic yeah i know i know i'm er he's so much you know everyone thinks yeah all he's also heavily for no sex before marriage really? yeah yeah he is like major he is he is he is he is he is he is really? so much he is instilled values oh my gosh like his friends yeah are so
    \item 161. you had to be married catholics you know catholics cannot oh catholics aren't allowed contraception no yeah can't have sex yes hello i'm here everybody else had to sort of produce their marriage certificate mm yeah but the ones who really needed it didn't get it then yeah mm no yeah but it really yes it's quite recent we're quite if you yeah well women have only
    \item 173. yeah three months it's gonna take us about that time oh for a child's passport now we need erm all of the grandparents'information we need everything it's ridiculous you need marriage certificates and all sorts it's is this because people are abducting kids do you think? are they clamping you know you know? i don't know but when kids get taken
    \item 174. and i think that's my that's my fear is that like with the i don't ever wanna be driven by security or by by s by protect by protecting my family or by marriage or by cos they're all good things but i don't ever want them to be my main thing right that's cool which is the change like to see jesus come in and change
\end{itemize}